\documentclass[preprint,12pt]{elsarticle}

\usepackage{lineno,hyperref}
\usepackage{amsmath,amssymb,amsthm}
\usepackage{booktabs}
\usepackage{array}
\usepackage{makecell}
\usepackage{graphicx}
\usepackage{multirow}
\usepackage{xcolor}
\usepackage{algorithm}
\usepackage{algpseudocode}
\usepackage{placeins}
\usepackage{rotating}
\usepackage[export]{adjustbox}
\usepackage{tikz}
\usetikzlibrary{arrows.meta,positioning,calc,backgrounds,shapes.geometric}
\definecolor{treeA}{HTML}{2A78D6}
\definecolor{compC}{HTML}{EB6834}
\definecolor{treeB}{HTML}{9A9994}
\definecolor{critF}{HTML}{FFE066}
\definecolor{critD}{HTML}{B8860B}
\newcommand{\fittable}[1]{\begin{adjustbox}{max width=\textwidth}#1\end{adjustbox}}
\newcommand{\tablenote}[1]{\par\smallskip\begin{minipage}{\textwidth}\footnotesize\linespread{1}\selectfont #1\end{minipage}}

\newtheorem{definition}{Definition}
\newtheorem{proposition}{Proposition}

\newcommand{\Gx}{G_x}
\newcommand{\comp}{c^{\dagger}}
\DeclareRobustCommand{\cfull}{\tikz[baseline=-0.55ex]\fill[black] (0,0) circle (0.6ex);}
\DeclareRobustCommand{\chalf}{\tikz[baseline=-0.55ex]{\fill[black] (0,0) -- (90:0.6ex) arc (90:270:0.6ex) -- cycle; \draw[black, line width=0.4pt] (0,0) circle (0.6ex);}}
\DeclareRobustCommand{\cempty}{\tikz[baseline=-0.55ex]\draw[black, line width=0.4pt] (0,0) circle (0.6ex);}

\newcommand{\nDatasets}{16}
\newcommand{\nRepetitions}{10}

\newcommand{\residualMax}{1.4\times 10^{-14}}
\newcommand{\faithfulCount}{342{,}040}
\newcommand{\critCoverageMin}{97.7}
\newcommand{\critCoverageMax}{100.0}
\newcommand{\nSamples}{342{,}040}

\newcommand{\interWinsMinAll}{12}
\newcommand{\interPMaxAll}{0.039}

\newcommand{\tweakReachMin}{73}
\newcommand{\tweakReachMax}{92}
\newcommand{\critReachMin}{18}
\newcommand{\critReachMax}{41}
\newcommand{\costRatioMin}{2.2}
\newcommand{\costRatioMax}{2.8}

\newcommand{\dpgMsMin}{87}
\newcommand{\dpgMsMax}{461}

\newcommand{\anchorsNDatasets}{14}
\newcommand{\dpgShapRatioMax}{11}
\newcommand{\slowRatioMin}{35}
\newcommand{\slowRatioMax}{504}
\newcommand{\dpgNodesMin}{173}
\newcommand{\dpgNodesMax}{885}
\newcommand{\cvPredMin}{26}
\newcommand{\cvPredMax}{123}

\newcommand{\eOneCritMin}{31}
\newcommand{\eOneCritMax}{40}

\newcommand{\eOneRandMin}{6}
\newcommand{\eOneRandMax}{8}

\newcommand{\eTwoTrueGainMin}{6}
\newcommand{\eTwoTrueGainMax}{13}

\newcommand{\eTwoNotCompMin}{18}
\newcommand{\eTwoNotCompMax}{31}
\newcommand{\pairBetterMin}{92}
\newcommand{\pairBetterMax}{97}

\newcommand{\eTwoErrors}{49{,}835}

\newcommand{\eTwoISepCritMin}{0.25}
\newcommand{\eTwoISepCritMax}{0.33}

\newcommand{\eTwoISepGuidedSig}{4}
\newcommand{\eTwoISepGuidedN}{8}

\newcommand{\cfSizeMin}{0.76}
\newcommand{\cfSizeMax}{1.01}
\newcommand{\shareMin}{1.05}
\newcommand{\shareMax}{28.7}
\newcommand{\shareRFMin}{1.05}
\newcommand{\shareRFMax}{2.37}

\newcommand{\shareRhoMin}{0.05}
\newcommand{\shareRhoMax}{0.47}

\newcommand{\noDigitsN}{15}
\newcommand{\fairFlipCritMin}{18}
\newcommand{\fairFlipCritMax}{44}
\newcommand{\fairFlipShapMin}{15}
\newcommand{\fairFlipShapMax}{37}
\newcommand{\fairFlipLimeMin}{12}
\newcommand{\fairFlipLimeMax}{32}

\newcommand{\fairFlipOracleMin}{46}
\newcommand{\fairFlipOracleMax}{75}
\newcommand{\fairHitCritMin}{46}
\newcommand{\fairHitCritMax}{69}
\newcommand{\fairHitShapMin}{29}
\newcommand{\fairHitShapMax}{50}

\newcommand{\fairSepCritMin}{0.26}
\newcommand{\fairSepCritMax}{0.34}
\newcommand{\fairSepOracleMin}{0.13}
\newcommand{\fairSepOracleMax}{0.18}
\newcommand{\fairAgreeCritMin}{31}
\newcommand{\fairAgreeCritMax}{40}
\newcommand{\fairAgreeOracleMin}{8}
\newcommand{\fairAgreeOracleMax}{18}
\newcommand{\fairSizeCritMin}{0.88}
\newcommand{\fairSizeCritMax}{1.34}
\newcommand{\fairSizeOracleMin}{1.62}
\newcommand{\fairSizeOracleMax}{2.13}

\newcommand{\fairTreesCritMin}{8}
\newcommand{\fairTreesCritMax}{27}
\newcommand{\fairTreesOracleMin}{2}
\newcommand{\fairTreesOracleMax}{7}
\newcommand{\atypCritMin}{15}
\newcommand{\atypCritMax}{22}
\newcommand{\atypCfMin}{17}
\newcommand{\atypCfMax}{18}

\newcommand{\atypSample}{5}
\newcommand{\gainRhoMin}{0.52}
\newcommand{\gainRhoMax}{0.68}

\newcommand{\fairFlipShapSig}{3}

\newcommand{\fairHitShapSig}{4}
\newcommand{\pairEstMin}{49}
\newcommand{\pairEstMax}{70}

\newcommand{\plFeatCritMin}{54}
\newcommand{\plFeatCritMax}{73}
\newcommand{\plFeatRandMin}{11}
\newcommand{\plFeatRandMax}{29}
\newcommand{\plFeatShapMin}{68}
\newcommand{\plFeatShapMax}{93}
\newcommand{\plThrCritMin}{46}
\newcommand{\plThrCritMax}{71}
\newcommand{\plThrRandMin}{6}
\newcommand{\plThrRandMax}{21}

\newcommand{\plThrShapNearMin}{5}
\newcommand{\plThrShapNearMax}{30}
\newcommand{\plThrShapEstMin}{59}
\newcommand{\plThrShapEstMax}{90}

\newcommand{\plReps}{10}
\newcommand{\eThreeDpgMin}{-0.6}
\newcommand{\eThreeDpgMax}{+0.5}
\newcommand{\eThreeAcceptMin}{0.1}
\newcommand{\eThreeAcceptMax}{0.8}
\newcommand{\eThreeRefitMin}{11}
\newcommand{\eThreeRefitMax}{15}
\newcommand{\eThreeRefitAda}{8.5}

\newcommand{\eThreeNTests}{42}

\newcommand{\exN}{10}
\newcommand{\exSameFeature}{5}
\newcommand{\exGainSame}{5}
\newcommand{\exGainBetter}{3}

\journal{Knowledge-Based Systems}

\begin{document}

\begin{frontmatter}

\title{Where Tree Ensembles Decide: Structural Local Explanation with Decision Predicate Graphs}

\author[trieste]{Sylvio Barbon Junior\corref{cor1}}
\ead{sylvio.barbonjunior@units.it}
\cortext[cor1]{Corresponding author.}

\author[cork]{Luca Longo}
\ead{LLongo@ucc.ie}

\address[trieste]{Department of Engineering and Architecture, University of Trieste,
Via Alfonso Valerio 6/1, 34127 Trieste, Italy}

\address[cork]{School of Computer Science and Information Technology, University College Cork, Cork, T12 XF62, Ireland}

\begin{abstract}
Tree ensembles are among the most accurate models for tabular data, yet the computation behind a
single prediction is spread over hundreds of decision rules. Feature attributions and counterfactuals
explain such predictions through the influence of features or a change of the input, leaving open
which of the thresholds applied by the trees a prediction responds to, while structural explanations
compress these rules and lose the trees behind them. To address this gap, we propose a local Decision
Predicate Graph (local DPG), which gathers the rules executed for a sample into a verifiable record of
the computation, with a class support that reproduces every explained prediction. From these rules we
select the critical predicate, the executed predicate whose crossing is estimated to move the model
most strongly toward a chosen class, which offers the user a concrete threshold and the trees that
share it, without an external explainer. Across \nDatasets{} datasets and four ensemble families, it
separates differently predicted neighbours better than random and nearest-threshold predicates. A
single crossing recovers the true class in \fairHitCritMin--\fairHitCritMax\% of wrong predictions,
against \fairHitShapMin--\fairHitShapMax\% when the same rule is applied to the feature ranked first by
TreeSHAP, a measure close to the selection criterion, and on planted faults it finds the faulty
threshold in \plThrCritMin--\plThrCritMax\% of the resulting errors, against
\plThrRandMin--\plThrRandMax\% at random. Although training-free edits of these predicates did not reduce
the error on unseen data, the critical predicate complements attributions and counterfactuals by
naming where a prediction can be changed.
\end{abstract}


\end{frontmatter}

\section{Introduction}
\label{sec:introduction}

Tree ensembles, such as random forests, bagging and gradient boosting, remain among the most
accurate and widely deployed models for tabular data. Their predictions result from the
aggregation of hundreds of decision trees, each of which routes the input through a sequence
of threshold tests, so the computation behind a single prediction is distributed across the
model. Local explanation methods aim to make this computation intelligible for one input at a
time~\citep{lundberg2020local,vilone2021notions,LONGO2024102301}, and they serve different
purposes, among them justifying a prediction, controlling what a model has learned, and
improving it through an understanding of its behaviour~\citep{adadi2018peeking}.
Improvement is the use that practitioners report most often, since explanations are mainly
employed to debug models~\citep{bhatt2020explainable,kaur2020interpreting}, and debugging begins
with understanding why individual predictions fail~\citep{nushi2018towards,kulesza2015principles}.
This paper addresses the first step of this purpose, the examination of individual predictions,
and asks, for a trained ensemble and a given input, which of the predicates executed by the model,
if crossed, would most change the prediction toward an alternative class.

A prediction of a tree ensemble can be explained from three complementary points of view.
Feature attributions such as SHapley Additive exPlanations (SHAP) and Local Interpretable
Model-agnostic Explanations (LIME) describe the \emph{influence} of each input,
quantifying how much it contributed to the
output~\citep{lundberg2017unified,lundberg2020local,ribeiro2016why}. Counterfactual explanations
move instead to the \emph{input space}, where they describe a change of the input that would
have produced a different
prediction~\citep{wachter2018counterfactual,tolomei2017interpretable,parmentier2021optimal,guidotti2019factual}. A third
view looks at the \emph{logical structure} of the model, that is, at the predicates and rules that
the trees actually executed, and underlies the route-based and formal explanations of tree
ensembles~\citep{hatwell2020chirps,mollas2022conclusive,izza2021explaining,arrighi2024decision}.
These perspectives are not interchangeable, and a growing body of
work shows how far they can diverge. A large attribution does not imply that a feature affects
the prediction~\citep{fernandezloria2022explaining}, the features ranked first by attribution
methods are often neither necessary nor sufficient to change
it~\citep{mothilal2021towards,watson2021local}, and Shapley values may even rank irrelevant
features above relevant ones~\citep{kumar2020problems,huang2024failings}. Counterfactuals, in
turn, indicate what should change in the input, yet they do not single out a predicate of the
model at which the prediction can be changed. This is the question that motivates the present work, and it is
the structural view that can answer it.

For a single decision tree, the logical structure of a prediction is easy to describe, since it
coincides with the root-to-leaf route followed by the sample, a conjunction of predicates of the form
$x_j \le \theta$ or $x_j > \theta$ that reads as a rule. A tree ensemble applies many such rules
at once, one per tree, some of which support the predicted class while others support an alternative
class. Existing structural explanations compress, select or aggregate these routes, and in doing so
they lose three elements of this computation, namely the complete set of rules that were executed,
the trees that share each predicate, and the rules that lead to competing classes. Preserving these
elements makes it possible to ask where, among the executed predicates, a prediction can be changed.
Our hypothesis is that, when the trees of an ensemble reuse the same split thresholds, crossing a single
predicate shared by a group of trees can change the prediction in a substantial share of cases, even
though the evidence for a competing class is spread over many predicates, and that this part of the
model's computation is made explicit neither by attributions nor by counterfactuals.

We therefore propose a \emph{structural local explanation} that follows the computation of the
ensemble for one sample. A \emph{local Decision Predicate Graph} (local DPG) gathers the executed
rules into a single graph, built only from the paths of this sample rather than from a graph
induced from the training data as in the original formulation of DPGs~\citep{arrighi2024decision},
and keeps on every node and edge the identity of the trees that executed it, while the class support
is computed from the original contributions of the trees. From these rules we select the
\emph{critical predicate}, the executed predicate whose threshold, if crossed, is estimated to move
the ensemble most strongly toward a chosen class, which addresses why the model predicted one class
rather than another~\citep{miller2019explanation}.

The contributions of this paper are organised in three levels, a structure that records the
computation behind a prediction, an insight drawn from this structure, and its use on wrong
predictions.
\begin{enumerate}
  \item We propose the local DPG, which represents every predicate executed for a sample, records the
    trees that executed it and derives a class support that reproduced every explained prediction,
    at a cost of \dpgMsMin{} to \dpgMsMax{}\,ms per sample (Sections~\ref{sec:method}
    and~\ref{sec:res-rq4}). The graph is too large to be read by a person, and we propose it as a
    verifiable record of the computation.
  \item We define the critical predicate, the element offered to the user, which requires no external
    explainer. It exceeds random and nearest-threshold predicates on measures that do not involve its
    selection, and its crossing, which changes a single feature, alters the prediction more often
    than the same rule applied to the feature ranked first by TreeSHAP or LIME in most families
    (Section~\ref{sec:res-rq1}).
  \item We show how the critical predicate helps to examine wrong predictions. A single crossing
    recovers the true class in \fairHitCritMin--\fairHitCritMax\% of cases, against
    \fairHitShapMin--\fairHitShapMax\% when the same rule is applied to the feature ranked first by
    contrastive TreeSHAP, and it locates faults planted in synthetic data
    (Section~\ref{sec:res-rq2}).
\end{enumerate}

The experiments cover \nDatasets{} datasets and four ensemble families and are organised around four
research questions, on how the critical predicate compares with other predicates (RQ1), wrong
predictions (RQ2), training-free repair (RQ3), and cost and size (RQ4). They also mark the limits of
this use, since training-free edits of the localised predicates do not reduce the error on unseen data
(Section~\ref{sec:res-rq3}).

Section~\ref{sec:related-work} reviews related work through a taxonomy based on these three views. Section~\ref{sec:method} formalises the
executed traces of an ensemble and presents the local DPG, the critical predicate and its
use. Section~\ref{sec:design} describes the experimental design, and
Section~\ref{sec:results} reports the results for the four research questions.
Section~\ref{sec:discussion} discusses implications, scope and limitations, and
Section~\ref{sec:conclusion} concludes the paper.

\section{Related work}
\label{sec:related-work}

\begin{figure}[!t]
\centering
\linespread{1}\selectfont
\begin{tikzpicture}[
  font=\sffamily\footnotesize,
  root/.style={draw=black!70, fill=black!8, rounded corners=2pt, rotate=90,
    align=center, inner sep=3pt, font=\sffamily\footnotesize\bfseries},
  view/.style={draw=black!60, fill=black!4, rounded corners=2pt, text width=1.55cm,
    align=center, inner sep=3pt, font=\sffamily\footnotesize\bfseries},
  scope/.style={draw=black!50, fill=white, rounded corners=2pt, text width=0.95cm,
    align=center, inner sep=2.5pt},
  leaf/.style={draw=black!35, fill=white, rounded corners=2pt, text width=8.3cm,
    align=left, inner sep=2.5pt, anchor=north west},
  ours/.style={leaf, draw=treeA, fill=treeA!10, line width=0.9pt},
  edge/.style={black!55, line width=0.6pt},
  next/.style={below=1.2mm of #1.south west, anchor=north west},
  group/.style={below=3.2mm of #1.south west, anchor=north west},
]
\def\lx{4.35}
\node[leaf] (l1) at (\lx, 0) {\textbf{Rule sets}\quad inTrees~\citep{deng2019interpreting}, SIRUS~\citep{benard2021sirus}, RuleFit~\citep{friedman2008predictive}};
\node[leaf, next=l1] (l2) {\textbf{Single-tree surrogates}\quad born-again trees~\citep{vidal2020born}, forest-based tree~\citep{sagi2020explainable}, explainable ensemble trees~\citep{aria2024explainable}};
\node[leaf, next=l2] (l3) {\textbf{Predicate graphs}\quad original DPG~\citep{arrighi2024decision}, its extension to isolation forests~\citep{ceschin2025extending}};
\node[leaf, group=l3] (l4) {\textbf{Route-based}\quad CHIRPS~\citep{hatwell2020chirps}, LionForests~\citep{mollas2022conclusive}, Bellatrex~\citep{dedja2023bellatrex}};
\node[leaf, next=l4] (l5) {\textbf{Formal}\quad abductive and majoritary reasons~\citep{izza2021explaining,audemard2022trading}, contrastive explanations~\citep{ignatiev2020contrastive}, sufficient and necessary reasons~\citep{darwiche2020reasons}};
\node[leaf, next=l5] (l6) {\textbf{Argumentation}\quad bipolar argumentation frameworks~\citep{potyka2023explaining}};
\node[ours, next=l6] (l7) {\textbf{\textcolor{treeA}{Local DPG (this work)}}\quad executed predicates, the trees that share them and the critical predicate};
\node[leaf, group=l7] (l8) {\textbf{Feature attribution}\quad TreeSHAP~\citep{lundberg2020local}, path contributions~\citep{palczewska2014interpreting}};
\node[leaf, next=l8] (l9) {\textbf{Local surrogates and rules}\quad LIME~\citep{ribeiro2016why}, Anchors~\citep{ribeiro2018anchors}, LORE~\citep{guidotti2019factual,guidotti2024stable}};
\node[leaf, group=l9] (l10) {\textbf{Counterfactuals and robustness}\quad Wachter et al.~\citep{wachter2018counterfactual}, feature tweaking~\citep{tolomei2017interpretable}, FOCUS~\citep{lucic2022focus}, optimal counterfactuals~\citep{parmentier2021optimal}, evasion~\citep{kantchelian2016evasion}, robustness verification~\citep{chen2019robustness}};
\def\sx{3.5}
\path let \p1=($(l1.north)!0.5!(l3.south)$) in node[scope] (sg) at (\sx,\y1) {Global};
\path let \p1=($(l4.north)!0.5!(l7.south)$) in node[scope] (sl) at (\sx,\y1) {Local};
\path let \p1=($(l8.north)!0.5!(l9.south)$) in node[scope] (si) at (\sx,\y1) {Local};
\path let \p1=($(l10.north)!0.5!(l10.south)$) in node[scope] (sc) at (\sx,\y1) {Local};
\def\vx{1.8}
\path let \p1=($(l1.north)!0.5!(l7.south)$) in node[view] (vs) at (\vx,\y1) {Logical structure};
\path let \p1=(si) in node[view] (vi) at (\vx,\y1) {Influence};
\path let \p1=(sc) in node[view] (vc) at (\vx,\y1) {Input space};
\path let \p1=($(vs)!0.5!(vc)$) in node[root] (r) at (0.25,\y1) {Explanations of tree ensembles};
\foreach \v in {vs, vi, vc} \draw[edge] (r.south) -- ++(0.25,0) |- (\v.west);
\foreach \s in {sg, sl} \draw[edge] (vs.east) -- ++(0.2,0) |- (\s.west);
\draw[edge] (vi.east) -- (si.west);
\draw[edge] (vc.east) -- (sc.west);
\foreach \l in {l1, l2, l3} \draw[edge] (sg.east) -- ++(0.15,0) |- (\l.west);
\foreach \l in {l4, l5, l6} \draw[edge] (sl.east) -- ++(0.15,0) |- (\l.west);
\draw[edge, draw=treeA, line width=0.9pt] (sl.east) -- ++(0.15,0) |- (l7.west);
\foreach \l in {l8, l9} \draw[edge] (si.east) -- ++(0.15,0) |- (\l.west);
\draw[edge] (sc.east) -- (l10.west);
\end{tikzpicture}
\caption{Taxonomy of the explanation methods for tree ensembles reviewed in
Section~\ref{sec:related-work}, by the view of a prediction they adopt and by their scope.}
\label{fig:taxonomy}
\end{figure}
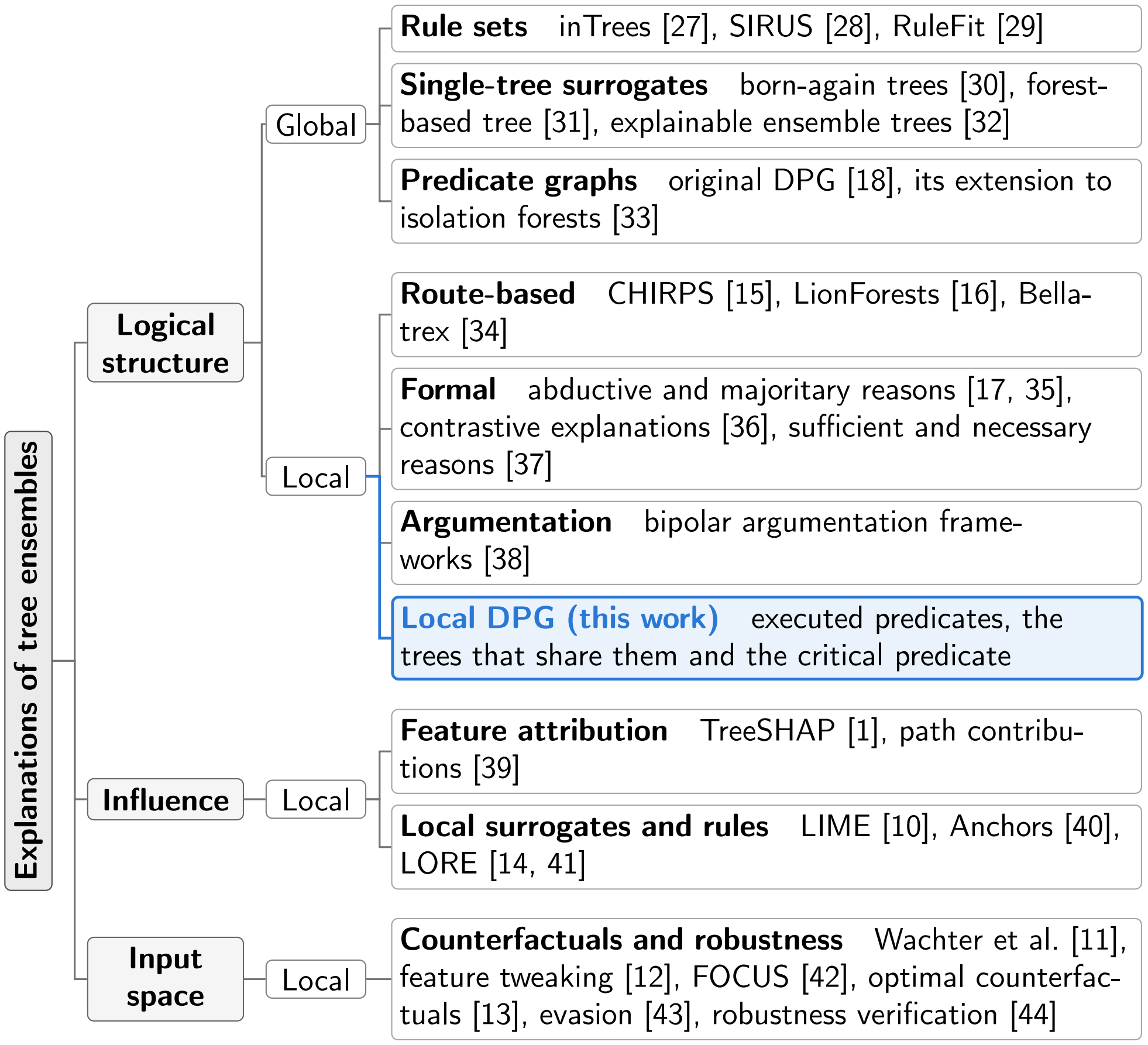

The introduction distinguished three views of a prediction, based on the influence of the features,
on a change of the input and on the logical structure of the model. We use these views to organise the
explanation methods for tree ensembles in the taxonomy of Figure~\ref{fig:taxonomy}, which cuts across
the usual classification by the kind of object a method returns~\citep{poeta2023concept,arrieta2020explainable}
and arranges the methods by the view they adopt and by their scope. It also locates the local DPG among
the local structural explanations. Table~\ref{tab:positioning} then compares the methods along the three
levels of this paper, namely whether the explanation returns the executed paths of the sample
(structure), which element it presents as decisive, in our reading of each method, and toward which
class (insight), and what is changed to alter the prediction (use).

\begin{table}[!tbp]
\centering
\caption{Positioning of explanation methods for tree ensembles, in the order of
Figure~\ref{fig:taxonomy}. Scope: local (L) or global (G). Access: model-direct (d) or
model-agnostic (a). Executed paths: complete, with the trees of each predicate (\cfull), or
summarised (\chalf). Target class: chosen, including the true class (\cfull), any other class
(\chalf), or per-class outputs without contrast (\cempty). Intervention: input change (I) or
threshold crossing (T). A dash marks an absent property.}
\label{tab:positioning}
\footnotesize
\linespread{1}\selectfont
\setlength{\tabcolsep}{3.5pt}
\begin{adjustbox}{max width=\textwidth}
\begin{tabular}{>{\raggedright\arraybackslash}p{4.6cm}cccl cc}
\toprule
 & & & \multicolumn{1}{c}{Structure} & \multicolumn{2}{c}{Insight} & \multicolumn{1}{c}{Use} \\
\cmidrule(lr){4-4}\cmidrule(lr){5-6}\cmidrule(lr){7-7}
Method & Scope & Access & \makecell{Executed\\paths} & \makecell[l]{Decisive\\element} & \makecell{Target\\class} & \makecell{Inter-\\vention} \\
\midrule
inTrees~\citep{deng2019interpreting}, SIRUS~\citep{benard2021sirus}, RuleFit~\citep{friedman2008predictive} & G & d & n/a & -- & -- & -- \\
Single-tree surrogates~\citep{vidal2020born,sagi2020explainable,aria2024explainable} & G & d & n/a & -- & -- & -- \\
Original DPG~\citep{arrighi2024decision,ceschin2025extending} & G & d & n/a & -- & -- & -- \\
\midrule
CHIRPS~\citep{hatwell2020chirps} & L & d & \chalf & rule & -- & -- \\
LionForests~\citep{mollas2022conclusive} & L & d & \chalf & rule & -- & -- \\
Bellatrex~\citep{dedja2023bellatrex} & L & d & \chalf & rules & -- & -- \\
Abductive and sufficient reasons~\citep{izza2021explaining,audemard2022trading,darwiche2020reasons} & L & d & -- & feature subset & -- & -- \\
Contrastive (formal)~\citep{ignatiev2020contrastive} & L & d & -- & feature subset & \chalf & -- \\
Bipolar argumentation~\citep{potyka2023explaining} & L & d & -- & arguments & \chalf & -- \\
\midrule
TreeSHAP~\citep{lundberg2020local}, path contributions~\citep{palczewska2014interpreting} & L & d & -- & feature & \cempty & -- \\
LIME~\citep{ribeiro2016why} & L & a & -- & feature & \cempty & -- \\
Anchors~\citep{ribeiro2018anchors} & L & a & -- & rule & -- & -- \\
LORE~\citep{guidotti2019factual} & L & a & -- & rule & \chalf & -- \\
\midrule
Wachter et al.~\citep{wachter2018counterfactual} & L & a & -- & input change & \cfull & I \\
Feature tweaking~\citep{tolomei2017interpretable}, optimal counterfactuals~\citep{parmentier2021optimal} & L & d & -- & input change & \cfull & I \\
FOCUS~\citep{lucic2022focus} & L & d & -- & input change & \chalf & I \\
Evasion and robustness verification~\citep{kantchelian2016evasion,chen2019robustness} & L & d & -- & input change & \chalf & I \\
\midrule
Local DPG (this work) & L & d & \cfull & executed predicate & \cfull & T \\
\bottomrule
\end{tabular}
\end{adjustbox}
\end{table}

\subsection{Structural explanations}
Because every root-to-leaf path of a tree is a conjunctive rule, the logical structure of an ensemble
can be made explicit as a set of rules. Several methods describe the ensemble as a whole. inTrees, the
Stable and Interpretable RUle Set (SIRUS) and RuleFit extract or recombine its
rules~\citep{deng2019interpreting,benard2021sirus,friedman2008predictive}, while others replace it with a
single tree~\citep{vidal2020born,sagi2020explainable,aria2024explainable}, or generate rules directly from the data \cite{LongoRizzo2024}. DPGs offer a graph-based
alternative, in which the nodes are the predicates of the splits, the edges are the transitions observed
along the paths of the training data, and the model is characterised through metrics such as
centrality~\citep{arrighi2024decision}, an idea later extended to isolation
forests~\citep{ceschin2025extending}. These methods describe the knowledge that an ensemble applies across
the data, and none of them shows which predicates produced a particular prediction.

Local structural methods start from the sample at hand. Collection of High Importance Random Path
Snippets (CHIRPS) condenses the patterns that recur among the executed paths into a single
rule~\citep{hatwell2020chirps}, LionForests selects a reduced set of paths and summarises them as feature
ranges~\citep{mollas2022conclusive}, and Bellatrex combines one representative rule per group of
trees~\citep{dedja2023bellatrex}. Selecting and compressing the routes discards their structure and the
paths that support competing classes, which is precisely the information a contrastive analysis requires.
Formal explanations bring logical guarantees instead. Abductive explanations identify a subset-minimal set
of feature values that entails the prediction, and can be computed for random forests with Boolean
satisfiability (SAT) solvers~\citep{izza2021explaining}, majoritary reasons relax minimality in exchange for
tractability~\citep{audemard2022trading}, and sufficient, necessary and complete reasons have been
formalised~\citep{darwiche2020reasons}. Their contrastive duals identify a minimal set of features whose
change can alter the prediction~\citep{ignatiev2020contrastive,izza2021explaining}. All of them take the
form of sets of conditions rather than routes and do not single out an executed predicate. Bipolar
argumentation frameworks, finally, represent the support and attack relations among the trees of a
forest~\citep{potyka2023explaining}, with arguments derived from the forest rather than from the predicates
it executed. The local DPG belongs to this group, building the graph for a single prediction and equipping
it with a class support and the critical predicate.

\subsection{Influence and local surrogates}
Feature attributions assign to each feature a contribution to the output. TreeSHAP computes SHAP values
exactly for tree models in polynomial time~\citep{lundberg2017unified,lundberg2020local}, and feature
contributions accumulate the change in class probability along each executed
path~\citep{palczewska2014interpreting}. Both summarise this information per feature, recording neither the
predicates that were executed nor the trees that supported a competing class. Several studies have
questioned how far such scores explain a prediction. Large weights need not affect the decision a model
makes~\citep{fernandezloria2022explaining}, the importance of a feature differs from its influence relative
to a reference value~\citep{framling2023importanceinfluence}, top-ranked features are often neither necessary
nor sufficient for a change of outcome~\citep{mothilal2021towards,watson2021local}, and Shapley values can
rank provably irrelevant features above relevant
ones~\citep{kumar2020problems,huang2024failings}. Our evaluation borrows this vocabulary only operationally,
asking whether the counterfactual found by feature tweaking crosses a predicate and whether the
single-feature crossing of the predicate changes the prediction.

Local surrogates describe the model around the sample. LIME fits a sparse linear
model~\citep{ribeiro2016why}, Anchors searches for a rule under which the prediction remains fixed with high
precision~\citep{ribeiro2018anchors}, and the LOcal Rule-based Explanations (LORE) method learns a decision
tree on a synthetic neighbourhood, from which it extracts factual and counterfactual
rules~\citep{guidotti2019factual,guidotti2024stable}. Being model-agnostic, they describe how the model
behaves near the sample, and their rules are not statements about the predicates the model itself executed.
Rule sets produced by such model-agnostic extractors have been shown to differ considerably when compared through quantitative metrics \citep{ViloneRL20}.

\subsection{Counterfactuals and robustness}
Feature tweaking considers every tree path that leads to the desired class, computes the smallest change
that satisfies it and retains the cheapest change that alters the prediction of the
ensemble~\citep{tolomei2017interpretable}. Flexible Optimizable CounterfactUal explanations for tree enSembles
(FOCUS) pursues the same goal through a differentiable approximation~\citep{lucic2022focus}, and mixed-integer
programming finds provably optimal counterfactuals~\citep{parmentier2021optimal}. The robustness of tree
ensembles raises a closely related question, for which \citet{kantchelian2016evasion} compute the nearest
instance that receives a different prediction and \citet{chen2019robustness} verify the minimal adversarial
perturbation or a guaranteed lower bound on it. These methods search the input space for a small change,
which may involve several features, and some offer guarantees on its size. The critical predicate asks a
narrower question, restricted to one feature and to the thresholds the trees executed for the sample, and it
returns an executed threshold and the trees that share it, without a guarantee of minimality.

\section{Proposed approach}
\label{sec:method}

\begin{figure}[!tbp]
\centering
\linespread{1}\selectfont
\resizebox{\textwidth}{!}{%
\begin{tikzpicture}[
  font=\sffamily\small,
  tn/.style={circle, draw=black!45, fill=white, inner sep=0pt, minimum size=2.8mm},
  te/.style={circle, draw=black!80, fill=black!80, inner sep=0pt, minimum size=2.8mm},
  lf/.style={rectangle, draw=black!40, fill=black!6, inner sep=0pt, minimum size=2.8mm},
  lp/.style={rectangle, draw=treeA, fill=treeA, inner sep=0pt, minimum size=3mm},
  lc/.style={rectangle, draw=compC, fill=compC, inner sep=0pt, minimum size=3mm},
  ge/.style={black!35, line width=0.5pt},
  xe/.style={black!80, line width=1.4pt},
  pred/.style={draw=black!60, rounded corners=3pt, fill=black!4, minimum height=6mm, inner sep=3pt, font=\small},
  crit/.style={pred, draw=critD, fill=critF, line width=1.6pt},
  halo/.style={critF, line width=5pt, line cap=round},
  cls/.style={draw=black!60, rounded corners=3pt, minimum height=6mm, minimum width=7mm, inner sep=3pt, font=\small, text=white},
  de/.style={-{Stealth[length=1.8mm]}, black!70, line width=0.8pt},
  dd/.style={de, densely dashed},
  tset/.style={font=\sffamily\scriptsize, text=black!65, fill=white, inner sep=0.8pt},
  flow/.style={-{Stealth[length=3mm]}, black!45, line width=1.3pt},
  title/.style={font=\sffamily\bfseries, align=center},
  sub/.style={font=\sffamily\footnotesize, align=center, text=black!65},
  band/.style={draw=black!15, fill=black!2, rounded corners=6pt},
  bandlabel/.style={font=\sffamily\footnotesize\bfseries, text=black!55, anchor=north west},
]
\newcommand{\minitree}[9]{%
  \begin{scope}[xshift=#1, scale=0.62, transform shape]
    \node[te] (#2r) at (0,0) {};
    \node[#3] (#2a) at (-0.6,-0.65) {};
    \node[#4] (#2b) at (0.6,-0.65) {};
    \draw[#5] (#2r) -- (#2a); \draw[#6] (#2r) -- (#2b);
    \foreach \p/\dx in {a/-0.6, b/0.6}{
      \node[lf] (#2\p1) at (\dx-0.3,-1.3) {}; \node[lf] (#2\p2) at (\dx+0.3,-1.3) {};
      \draw[ge] (#2\p) -- (#2\p1); \draw[ge] (#2\p) -- (#2\p2);
    }
    \node[#9] at (#2#7#8) {}; \draw[xe] (#2#7) -- (#2#7#8);
    \node[font=\sffamily\small, text=black!60] at (0,-1.95) {$T_{#2}$};
  \end{scope}}

\node[title] at (3.6,0.75) {(a) Tree ensemble};
\minitree{0.7cm}{1}{te}{tn}{xe}{ge}{a}{1}{lp}
\minitree{2.15cm}{2}{te}{tn}{xe}{ge}{a}{2}{lc}
\minitree{3.6cm}{3}{tn}{te}{ge}{xe}{b}{1}{lp}
\minitree{5.05cm}{4}{tn}{te}{ge}{xe}{b}{2}{lc}
\minitree{6.5cm}{5}{te}{tn}{xe}{ge}{a}{2}{lp}
\node[sub] at (3.6,-1.75) {each tree executes one root-to-leaf route for $x$};
\draw[flow] (7.05,-0.5) -- (7.75,-0.5);

\node[title] at (11.75,0.75) {(b) Executed traces and class support};
\begin{scope}[font=\footnotesize]
  \node[anchor=west] at (7.85,0.05)  {$\pi_1$: $x_3{\le}2.4 \to x_7{>}{-0.6} \to x_5{>}0.7 \to x_1{\le}3.2 \to$ \textcolor{treeA}{$\blacksquare$}\,$c_1$};
  \node[anchor=west] at (7.85,-0.45) {$\pi_2$: $x_3{\le}2.4 \to x_7{>}{-0.6} \to x_5{>}0.7 \to x_1{>}1.8 \to$ \textcolor{compC}{$\blacksquare$}\,$c_2$};
  \node[anchor=west] at (7.85,-0.95) {$\pi_3$, $\pi_5$: $x_3{\le}2.4 \to x_7{>}{-0.6} \to$ \colorbox{critF}{\strut$x_8{\le}0.2$} $\to$ \textcolor{treeA}{$\blacksquare$}\,$c_1$};
  \node[anchor=west] at (7.85,-1.45) {$\pi_4$: $x_3{\le}2.4 \to x_7{>}{-0.6} \to x_9{>}1.4 \to x_2{\le}0.5 \to$ \textcolor{compC}{$\blacksquare$}\,$c_2$};
  \node[draw=black!30, rounded corners=2pt, fill=white, align=center, inner sep=3pt] (sup) at (11.75,-2.45)
       {class support $S_x = g\bigl(z_0 + \sum_b v_b(x)\bigr)$\\$S_x(c_1) > S_x(c_2)$};
\end{scope}

\node[title] at (4.2,-3.6) {(c) Local DPG};
\node[pred, fill=white] (src) at (7.5,-3.6) {$x$};
\node[pred] (n3) at (7.5,-4.6) {$x_3 \le 2.4$};
\node[pred] (n7) at (7.5,-5.7) {$x_7 > -0.6$};
\node[crit] (n8) at (4.4,-6.95) {$x_8 \le 0.2$};
\node[font=\sffamily\footnotesize\bfseries, text=critD, align=center] (critlab) at (2.0,-8.3) {critical predicate\\$p^{\ast}(c_2)$};
\draw[-{Stealth[length=2mm]}, critD, line width=1pt] (critlab.north) to[bend left=20] (n8.west);
\node[pred] (n5) at (7.5,-6.95) {$x_5 > 0.7$};
\node[pred] (n9) at (11.0,-6.95) {$x_9 > 1.4$};
\node[pred] (n1a) at (6.2,-8.2) {$x_1 \le 3.2$};
\node[pred] (n1b) at (8.8,-8.2) {$x_1 > 1.8$};
\node[pred] (n2) at (11.0,-8.2) {$x_2 \le 0.5$};
\node[cls, fill=treeA] (cy) at (5.3,-9.5) {$c_1$};
\node[cls, fill=compC] (cc) at (9.9,-9.5) {$c_2$};
\node[sub, anchor=east] at (cy.west) {$S_x(c_1)$\,};
\node[sub, anchor=west] at (cc.east) {\,$S_x(c_2)$};
\draw[de] (src) -- node[tset, right] {$T_{1\text{--}5}$} (n3);
\draw[de] (n3) -- node[tset, right] {$T_{1\text{--}5}$} (n7);
\begin{scope}[on background layer]\draw[halo] (n7) -- (n8) -- (cy);\end{scope}
\draw[de] (n7) -- node[tset, above, sloped] {$T_3,T_5$} (n8);
\draw[de] (n7) -- node[tset, right] {$T_1,T_2$} (n5);
\draw[de] (n7) -- node[tset, above, sloped] {$T_4$} (n9);
\draw[de] (n5) -- node[tset, left, pos=0.45] {$T_1$} (n1a);
\draw[de] (n5) -- node[tset, right, pos=0.45] {$T_2$} (n1b);
\draw[de] (n9) -- node[tset, right] {$T_4$} (n2);
\draw[de] (n8) -- node[tset, below, sloped] {$T_3,T_5$} (cy);
\draw[de] (n1a) -- node[tset, right] {$T_1$} (cy);
\draw[de] (n1b) -- node[tset, left] {$T_2$} (cc);
\draw[de] (n2) -- node[tset, right] {$T_4$} (cc);
\draw[flow, rounded corners=5pt] (sup.south) -- (11.75,-3.6) -- (src.east);
\node[sub, align=left, anchor=west, text width=3.4cm] at (12.3,-6.6)
     {every route is an executed rule and every edge records the trees that executed it; class
      support comes from the tree contributions, not from counting routes};

\begin{scope}[yshift=-10.75cm]
  \node[title] at (3.4,-0.05) {(d) Ranking toward a target class $c$};
  \foreach \w/\y/\c/\l/\t in {
      3.4/-0.85/critF/{x_8 \le 0.2}/{T_3,T_5},
      1.9/-1.45/black!25/{x_5 > 0.7}/{T_1},
      1.3/-2.05/black!25/{x_1 \le 3.2}/{T_1},
      0.8/-2.65/black!25/{x_3 \le 2.4}/{T_1,T_3,T_5}}{
    \node[anchor=east, font=\small] at (1.6,\y) {$\l$};
    \fill[\c] (1.75,\y-0.17) rectangle (1.75+\w,\y+0.17);
    \node[anchor=west, font=\sffamily\scriptsize, text=black!60] at (1.8+\w,\y) {$\t$};
  }
  \draw[black!50, line width=0.6pt] (1.75,-0.55) -- (1.75,-2.95);
  \node[anchor=west, font=\sffamily\footnotesize\bfseries] at (1.8,-0.85) {$p^{\ast}(c)$};
  \node[sub, font=\sffamily\scriptsize] at (3.5,-3.6) {gains add up over the trees that share a predicate, and\\
    crossing $x_8 \le 0.2$ sends $T_3$ and $T_5$ (now at $c_1$)\\
    to branches whose leaves support $c$};
\end{scope}

\begin{scope}[yshift=-10.75cm]
  \node[title] at (11.9,-0.05) {(e) Threshold crossing in the whole model};
  \node[pred, draw=critD, fill=critF, line width=1.6pt, align=center, font=\footnotesize, minimum width=3.3cm] (cross) at (10.0,-1.0)
       {$x_8 = 0.1 \;\to\; 0.2^{+}$\\[-1pt]{\scriptsize threshold crossing}};
  \node[pred, fill=white, align=center, font=\footnotesize, minimum width=3.3cm] (eval) at (10.0,-2.2)
       {re-evaluate all trees\\[-1pt]{\scriptsize routes testing $x_8$ in $(0.1, 0.2]$ change}};
  \node[pred, fill=white, align=center, font=\footnotesize, minimum width=3.3cm] (out) at (10.0,-3.4)
       {$\Delta P(c) = S_{x'}(c) - S_x(c)$\\[-1pt]{\scriptsize flip if the model now predicts $c$}};
  \draw[de] (cross) -- (eval); \draw[de] (eval) -- (out);
  \node[anchor=north west, align=left, text width=4.4cm, font=\sffamily\footnotesize] at (12.35,-0.6)
       {With $c = c_2$, the top competitor, a flip shows where the prediction is contested.\\[3pt]
        With $c = y$, the true class, it tests whether a wrong prediction can be recovered.\\[3pt]
        {\color{black!65} The explanation names the threshold ($x_8 = 0.2$) and the trees involved
        ($T_3$, $T_5$).}};
\end{scope}
\end{tikzpicture}}
\caption{Overview of the approach for a sample $x$ with predicted class $\hat y = c_1$ and top
competitor $c^{\dagger} = c_2$. (a) Each tree executes one rule. (b) Executed traces and class
support computed from the tree contributions. (c) Local DPG, whose routes are the executed traces.
(d) Critical predicate toward a target class $c$, highlighted in every panel. (e) Single-feature
threshold crossing evaluated in the whole model.}
\label{fig:overview}
\end{figure}
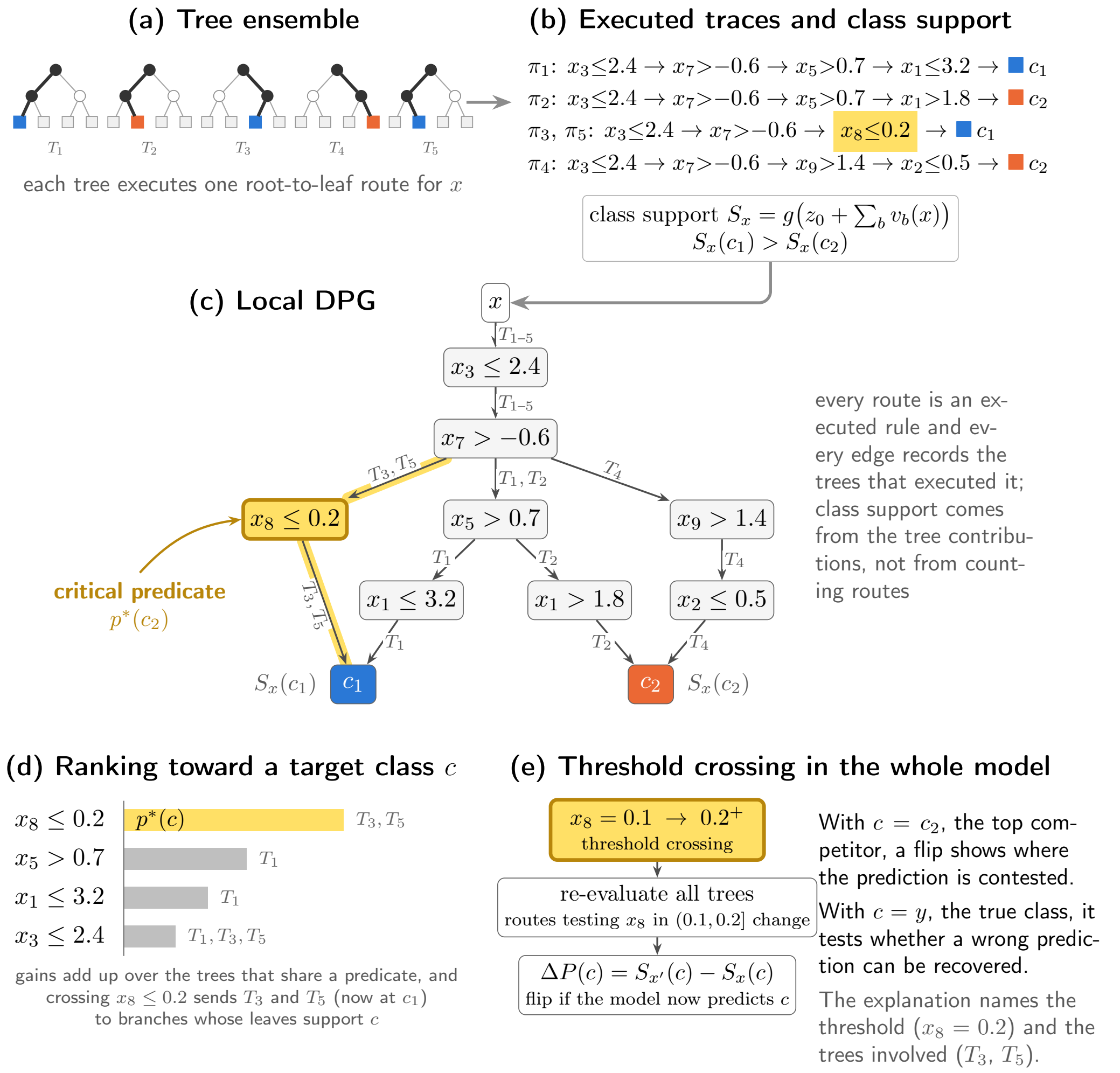

The approach follows the computation of the ensemble for one sample in five steps, illustrated
in Figure~\ref{fig:overview}. First, each tree executes one root-to-leaf rule for the sample
(panel a). Second, the support of each class is computed from the original contributions of the
trees, so that it reproduces the output of the ensemble (panel b). Third, the rules are gathered
into a local graph that keeps, on every node and edge, the identity of the trees that executed it
(panel c). Fourth, among the executed predicates, we select the one whose threshold, if
crossed, is estimated to move the ensemble most strongly toward a chosen class (panel d). Fifth,
the feature of that predicate is moved just past its threshold and the whole ensemble is
evaluated again, which shows how the prediction responds (panel e).

\subsection{Executed traces and class support}
\label{sec:traces}
\label{sec:support}

A tree ensemble predicts by sending the sample through every tree and combining the contributions of
the leaves it reaches (the notation is summarised in Table~\ref{tab:notation}, \ref{app:notation}). Let
$\mathcal{C} = \{c_1, \dots, c_K\}$ denote the set of classes and $\mathcal{F} = \{T_1, \dots, T_B\}$ a
fitted ensemble of $B$ decision trees over $d$ numeric features, a sample being
$x = (x_1, \dots, x_d) \in \mathbb{R}^d$. Each internal node compares one feature with a threshold and
sends the sample to its left child when $x_j \le \theta$ and to its right child otherwise. In tree $T_b$
the sample follows a sequence of such tests from the root to a leaf. Each test, read together with its
outcome, is an executed \emph{predicate} $p = (j, \theta, \circ)$, with $\circ \in \{\le, >\}$, and the
resulting sequence
\[
\pi_b(x) = \bigl(p^{(b)}_1, \dots, p^{(b)}_{L_b}, \; o_b(x)\bigr)
\]
is the \emph{trace} of $x$ in $T_b$, in which $L_b$ is the length of the path and $o_b(x) \in \mathcal{C}$
the class with which the trace is associated. The contribution of the tree may instead be spread over
several classes, as for a leaf of a random forest that holds 60\% of class $c_1$ and 40\% of class $c_2$,
or may be a signed score, as in gradient boosting. Class support is therefore computed from these
contributions and never by counting the routes that end in each class.

Predicates are compared through labels $\lambda(p) = $ \textquotedblleft$x_j \circ \theta\,$\textquotedblright,
in which the threshold is rounded to six decimal places, and predicates with the same label are treated as
one predicate. This grouping is a convention of the representation rather than an identity between
thresholds, and every crossing of a predicate uses the exact threshold of the tree concerned.

All the ensembles considered here compute their output by adding per-tree contributions $v_b(x)$. The
resulting state
\begin{equation}
z(x) = z_0 + \sum_{b=1}^{B} v_b(x)
\label{eq:state}
\end{equation}
is mapped by a link function $g$ to the class support $S_x = g(z(x))$, with $\sum_c S_x(c) = 1$, and
\ref{app:families} gives $z_0$, $v_b$, $g$ and $o_b$ for each family. The predicted class is
$\hat{y}(x) = \arg\max_c S_x(c)$, and we call $\comp(x) = \arg\max_{c \neq \hat{y}} S_x(c)$ the \emph{top
competitor}, breaking ties by class order. When the true class $y$ of a sample is known, the prediction is
\emph{wrong} if $\hat{y}(x) \neq y$, and in multiclass problems $y$ need not coincide with $\comp(x)$.

Statistics of the original DPG~\citep{arrighi2024decision} summarise how the training data traverse the
graph, and a ranking of classes obtained from them need not agree with the model. Defining the support
through Equation~\eqref{eq:state} avoids this mismatch.

\begin{proposition}[Output fidelity]
\label{prop:fidelity}
For every sample $x$, $\arg\max_c S_x(c)$ is the class predicted by the model, and the
contributions of the individual traces add up to the model's state,
$z_0 + \sum_b v_b(x) = z(x)$.
\end{proposition}

\begin{proof}
For each family, $v_b$ is precisely the quantity that the model sums and $g$ is the
model's own link function (\ref{app:families}). Hence $S_x$ equals the
model's predicted probabilities or, in the case of AdaBoost, a vote share whose
maximiser is the model's prediction.
\end{proof}

The proposition concerns the numbers attached to the explanation and not the topology of the graph, and
it holds up to floating-point rounding. Numerically, the explained class matched the prediction in
\faithfulCount{} of \nSamples{} cases, with a largest absolute residual of $\residualMax$.

\subsection{Local DPG}
\label{sec:route-faithfulness}

Each trace $\pi_b(x)$ defines a decision rule whose antecedent is the conjunction of its predicates and
whose consequent is its outcome. The local DPG $\Gx$ represents the rules executed for $x$ as a single
directed acyclic graph, with a source node, predicate nodes and one sink node for each class. Each trace
induces a path from the source through its predicates to the sink of its outcome, and every node and edge
is labelled with the set of trees whose traces traverse it. Merging predicates on their labels alone could
create source-to-sink paths that no tree executed, by joining the beginning of one trace to the end of
another. A predicate node is therefore identified by its label together with the labels that immediately
precede it on the trace, and the number of preceding labels is the smallest for which every source-to-sink
path, or \emph{route}, coincides with the label sequence of an executed trace. This condition is checked
exactly on the traces of the sample, and keeping the whole prefix of every trace always satisfies it.
Unlike the original DPG, which is induced from the training data, $\Gx$ is induced from the traces of $x$
alone, so that its routes enumerate the decision rules the model applied to $x$ and its tree sets identify
the trees that applied them (Figure~\ref{fig:overview}c).

\subsection{Critical predicate}
\label{sec:contrastive}

Explanations are often sought in contrastive form, asking why one outcome occurred rather than
another~\citep{miller2019explanation}. Intuitively, the critical predicate is the executed predicate whose
threshold, if crossed, is estimated to move the ensemble most strongly toward a chosen target class $c$.

To cross an executed predicate $p = (j, \theta, \circ)$ is to move $x_j$ to the other side of $\theta$. The
\emph{crossing value} $\tilde{x}_j$ is the value closest to $\theta$ on that side, in the precision in which
the trees compare features, and the \emph{slack} of the predicate, the margin by which the sample satisfies
its inequality, is $|x_j - \theta|$. Crossing $p$ in tree $T_b$ only, with all other trees left unchanged,
replaces $v_b(x)$ with the contribution $v_b(x^{(b,\ell)})$ of the leaf that $T_b$ reaches once $x_j$ is set
to the crossing value of its own threshold, where $\ell = \lambda(p)$.

\begin{definition}[Pivot predicate and critical predicate]
For a label $\ell$, let $\mathcal{B}_\ell$ be the set of trees whose traces execute a
predicate with label $\ell$. The \emph{estimated gain} of $\ell$ toward a target class
$c$ is
\begin{equation}
\Delta_\ell(c) = \sum_{b \in \mathcal{B}_\ell}
\Bigl[ g\bigl(z(x) - v_b(x) + v_b(x^{(b,\ell)})\bigr)_{c} - S_x(c) \Bigr].
\label{eq:gain}
\end{equation}
A label with $\Delta_\ell(c) > 0$ is a \emph{pivot predicate} toward $c$, and the pivot with the
largest gain is the \emph{critical predicate} $p^\ast(c)$, with ties broken first by smaller slack
and then by label. If no label has a positive gain, no critical predicate is returned. We write
$p^\ast$ for $p^\ast(\comp)$.
\end{definition}

The gain in Equation~\eqref{eq:gain} is a first-order estimate, in the sense that it adds the changes
obtained by crossing the predicate in each tree separately; it is not a derivative, and it ignores the
interaction between trees that are crossed together. It is computed directly from the stored traces and requires no
search over the input space, while the graph is the representation in which the selected predicate, its
trees and the competing routes can be inspected. A critical predicate toward the top competitor existed in
\critCoverageMin\% to \critCoverageMax\% of the samples, depending on the family. Because the gain sums over
the trees that execute the same predicate, the critical predicate tends to be shared by groups of trees, and it
usually lies on routes that reach the predicted class, since only the trees that do not yet support the
target can move toward it, as in Figure~\ref{fig:overview}, where $x_8 \le 0.2$ redirects $T_3$ and $T_5$. The
local DPG also lists the pivot predicates ranked by their gain, and the \emph{compact contrastive view}
retains the shortest prefix of this ranking that accounts for 80\% of the total positive gain.

\subsection{Using the critical predicate}
\label{sec:usage}

The critical predicate is used by intervening on the whole model. Given a target class $c$, the
\emph{threshold crossing} of $p^\ast(c)$ changes a single feature of the sample, setting $x_j$ to the crossing
value of the threshold nearest to the sample among the trees that share the label, and evaluates every tree
again on the crossed sample $x'$. We record the change in the support of the target class,
$\Delta P(c) = S_{x'}(c) - S_x(c)$, and whether the predicted class changes, which we call a \emph{flip}. This
\emph{whole-model intervention} changes one feature, not one node of the ensemble, so any tree whose route
tests $x_j$ between the old and the new value also changes its route, whether or not it executed the
predicate. It plays no role in the selection of $p^\ast(c)$, and throughout the paper crossing a predicate
refers to this intervention.

The target determines the question that the intervention answers. With the top competitor, $c = \comp$, the
critical predicate indicates where the prediction is estimated to be most contested. When the true class $y$ of
a wrongly predicted sample is known, the target $c = y$ asks whether a single threshold crossing turns the
prediction into the true class, and we say that the predicate \emph{recovers} the true class when the model then
predicts $y$. The two targets differ in multiclass problems whenever the true class is not the strongest
competitor.

\subsection{Algorithm and complexity}

Algorithm~\ref{alg:local-dpg} summarises the construction. With $L = \sum_b L_b$ the total length of the
executed paths and $D$ the maximum depth of the trees, extracting the traces takes $O(L)$ time in addition to
one evaluation of the ensemble, and estimating the gains takes $O(LD)$ time, since each executed predicate
re-evaluates the single tree that executed it. Choosing the context of the predicate nodes depends on how many
lengths are checked, for which we state no bound, and the measured runtimes of Section~\ref{sec:res-rq4} include
this step. The graph has at most $L + K + 1$ nodes and $L + B$ edges.

\begin{algorithm}[t]
\caption{Local DPG and contrastive view for a sample $x$.}
\label{alg:local-dpg}
\begin{algorithmic}[1]
\Require fitted ensemble $\mathcal{F}$, sample $x$, target class $c$ (by default $\comp$)
\For{$b = 1, \dots, B$}
  \State $\pi_b \gets$ executed trace of $x$ in $T_b$ \Comment{tree index stored}
  \State $v_b \gets$ contribution of $T_b$ at $x$
\EndFor
\State $z \gets z_0 + \sum_b v_b$; \; $S_x \gets g(z)$; \; $\hat y \gets \arg\max S_x$; \; $\comp \gets \arg\max_{c \ne \hat y} S_x$
\State $\Gx \gets$ merge of $\{\pi_b\}$, each predicate with the shortest preceding context for which every route is a trace
\State annotate each node and edge of $\Gx$ with its trees
\For{each executed predicate $(b, p)$}
  \State $\tilde x \gets x$ with $x_j$ set to the crossing value of $p$ in $T_b$
  \State $\Delta_{\lambda(p)}(c) \gets \Delta_{\lambda(p)}(c) + g(z - v_b + v_b(\tilde x))_{c} - S_x(c)$ \Comment{Eq.~\eqref{eq:gain}}
\EndFor
\State $p^\ast(c) \gets \arg\max_\ell \Delta_\ell(c)$ if $\max_\ell \Delta_\ell(c) > 0$ \Comment{ties: slack, then label}
\State \Return $\Gx$, $S_x$, $\comp$, ranked pivots, $p^\ast(c)$
\end{algorithmic}
\end{algorithm}

\section{Experimental design}
\label{sec:design}

The experiments address four research questions:
\begin{description}
  \item[RQ1 (critical predicate).] Compared with predicates selected at random, by a structural
    criterion, or from the feature ranked first by an attribution or counterfactual explainer, how
    does the critical predicate perform on measures that do not involve its selection, and does its
    estimated gain carry over to the whole model?
  \item[RQ2 (wrong predictions).] Directed toward the true class of a wrong prediction, how does the
    critical predicate perform on measures that do not involve its selection, how often does crossing
    it recover that class, and does it locate a fault planted in the training data?
  \item[RQ3 (repair).] Do the predicates identified on wrong predictions help training-free edits
    of the model reduce its error on unseen data?
  \item[RQ4 (cost and size).] What are the runtime and the size of the local DPG and of its compact
    contrastive view?
\end{description}

\subsection{Datasets and models}

We use \nDatasets{} public numeric classification datasets, obtained from
OpenML~\citep{vanschoren2014openml} and scikit-learn and most of them originating in the UCI
repository~\citep{dua2019uci}, with 4 to 617 features and 2 to 26 classes (Table~S15 of the supplementary
material). Each dataset is divided once, by stratified sampling, into a held-out test set of 20\% of the
samples and a remainder, which is split in the same way into a training core and a validation set in the
proportion 80:20. We study four ensemble families of 100 estimators, random forests (RF), bagging of decision
trees (Bagging), AdaBoost with Stagewise Additive Modeling using a Multi-class Exponential loss, SAMME
(AdaBoost) and gradient boosting (GBM). For each combination of dataset, family and repetition, a single
configuration is chosen on validation accuracy alone, among maximum depths of 4, 8 and unlimited for RF and
Bagging, 1, 2 and 3 for AdaBoost, and 2, 3 and 5 for GBM with a learning rate of $0.1$, and refitted on the
training and validation data. This refitted model is the one explained by every method and control, no explanation metric enters the
selection, the local DPG has no hyper-parameter to tune, and the
\nRepetitions{} repetitions differ in the random initialisation of the models and of the random controls. The code,
the fixed splits and the per-sample results are archived on Zenodo~\citep{barbon2026localdpg}.

\subsection{Explanation methods and controls}

The baselines are TreeSHAP~\citep{lundberg2020local} in its path-dependent form, LIME~\citep{ribeiro2016why},
Anchors~\citep{ribeiro2018anchors}, LORE~\citep{guidotti2019factual,guidotti2024stable} and feature
tweaking~\citep{tolomei2017interpretable}, which returns the cheapest change of the sample, in standardised
Euclidean distance, that satisfies the path to a leaf of the target class and changes the prediction of the
whole model. Because of their cost, Anchors and LORE explain the first 50 test samples in the first three
repetitions, and Anchors is reported for the \anchorsNDatasets{} datasets within the memory limit,
whereas all other methods explain every test sample in all \nRepetitions{} repetitions (Section~S1 of the
supplementary material).

\begin{figure}[!tbp]
\centering
\linespread{1}\selectfont
\resizebox{\textwidth}{!}{%
\begin{tikzpicture}[
  font=\sffamily\small,
  axis/.style={black!70, line width=0.8pt},
  exe/.style={black!85, line width=1.4pt},
  oth/.style={black!30, line width=1.0pt},
  smp/.style={diamond, fill=black, inner sep=0pt, minimum size=2.8mm},
  arr/.style={-{Stealth[length=2.4mm]}, line width=1.6pt},
  lab/.style={font=\sffamily\footnotesize, text=black!75},
  title/.style={font=\sffamily\bfseries, align=center},
  rule/.style={anchor=west, font=\sffamily\small, align=left},
]
\node[title] at (3.2,1.2) {(a) Predicates on the axis of each feature};
\draw[axis] (0.4,0) -- (6.4,0) node[right, font=\normalsize] {$x_8$};
\draw[exe] (3.0,-0.18) -- (3.0,0.18) node[above, lab] {$0.2$ ($T_3, T_5$)};
\node[smp] at (2.0,0) {};
\draw[arr, critD] (2.0,-0.38) -- (3.2,-0.38);
\begin{scope}[yshift=-1.5cm]
  \draw[axis] (0.4,0) -- (6.4,0) node[right, font=\normalsize] {$x_5$};
  \draw[exe] (1.3,-0.18) -- (1.3,0.18) node[above, lab] {$0.7$ ($T_1, T_2$)};
  \draw[exe] (5.4,-0.18) -- (5.4,0.18) node[above, lab] {$1.6$ ($T_6$)};
  \foreach \v in {3.2, 3.9, 4.6} \draw[oth] (\v,-0.14) -- (\v,0.14);
  \node[smp] at (2.3,0) {};
  \draw[arr, treeA] (2.3,-0.38) -- (1.1,-0.38);
  \draw[arr, black!45, densely dashed] (2.3,-0.65) -- (5.6,-0.65);
\end{scope}
\begin{scope}[yshift=-3.25cm]
  \draw[axis] (0.4,0) -- (6.4,0) node[right, font=\normalsize] {$x_1$};
  \draw[exe] (1.6,-0.18) -- (1.6,0.18) node[above, lab] {$1.8$};
  \draw[exe] (3.6,-0.18) -- (3.6,0.18) node[above, lab] {$2.55$ ($T_7$)};
  \draw[exe] (5.4,-0.18) -- (5.4,0.18) node[above, lab] {$3.2$ ($T_1$)};
  \node[smp] at (3.35,0) {};
  \draw[arr, black] (3.35,-0.38) -- (3.8,-0.38);
\end{scope}
\node[lab, align=left, anchor=north west] at (0.4,-3.95) {$\blacklozenge$ value of the sample\\
  dark ticks: executed predicates\\light ticks: thresholds of other trees};

\node[title] at (11.9,1.2) {(b) Predicate selected by each criterion};
\begin{scope}[xshift=8.8cm]
  \draw[arr, critD] (0,0.1) -- (0.7,0.1);
  \node[rule] at (0.85,0.1) {\textbf{Critical}: largest estimated gain,\\$x_8 \le 0.2$, shared by $T_3$ and $T_5$};
  \draw[arr, treeA] (0,-1.15) -- (0.7,-1.15);
  \node[rule] at (0.85,-1.15) {\textbf{TreeSHAP-guided}: feature ranked first\\($x_5$), predicate nearest to the sample};
  \draw[arr, black] (0,-2.4) -- (0.7,-2.4);
  \node[rule] at (0.85,-2.4) {\textbf{Nearest}: smallest slack among all\\executed predicates ($x_1 \le 2.55$)};
  \draw[arr, black!45, densely dashed] (0,-3.65) -- (0.7,-3.65);
  \node[rule] at (0.85,-3.65) {\textbf{Random}: any executed predicate, often\\far from the sample, past other thresholds};
\end{scope}
\end{tikzpicture}}
\caption{Predicate selected for crossing by the critical-predicate criterion and by three
controls, on a schematic sample. (a) Feature axes with the value of the sample, the executed
predicates (dark) and the thresholds of other trees (light). (b) Selection rule of each
criterion.}
\label{fig:controls}
\end{figure}
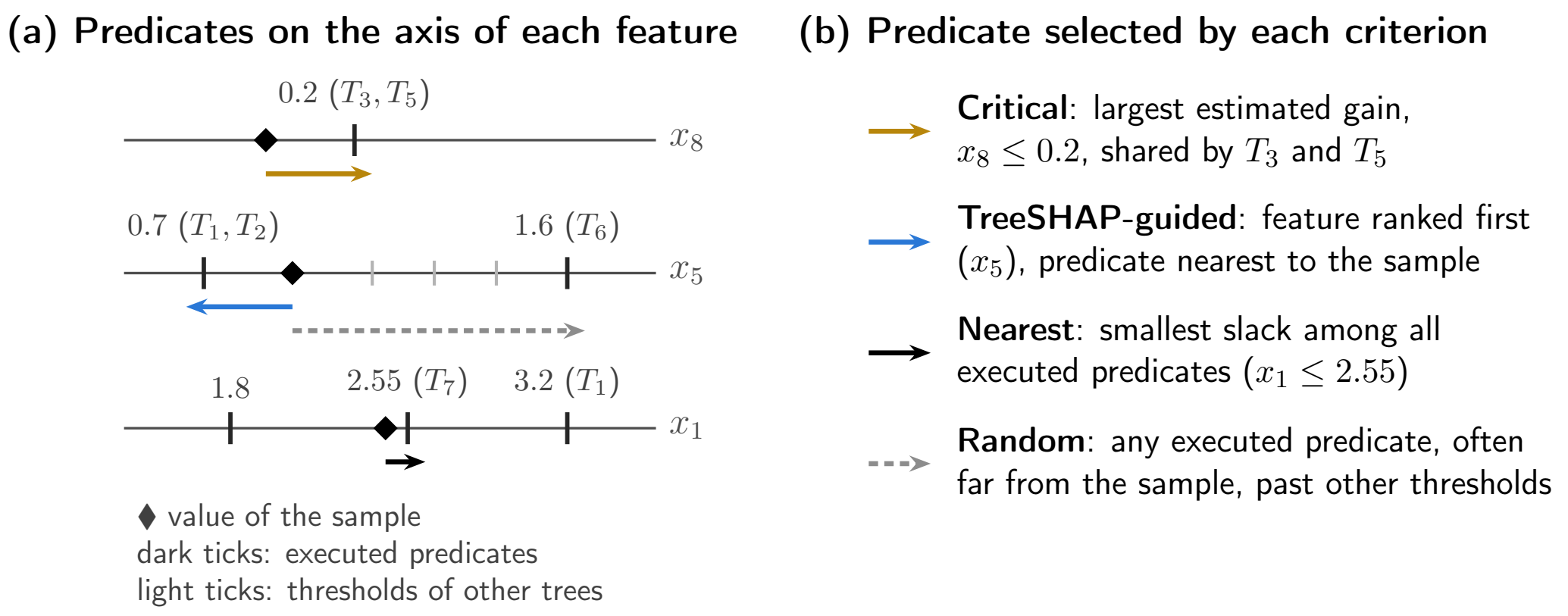

Each control selects a single executed predicate of the sample, which is then crossed exactly as the critical
predicate is (Figure~\ref{fig:controls}). Three controls are structural, namely a random executed predicate,
the predicate with the smallest slack (\emph{nearest}) and the predicate executed by the largest number of
trees (\emph{frequent}). The other five are guided by an explainer. Explainers score features or propose input
changes but do not return a predicate, so each control takes the feature that its explainer ranks first
(Section~S1 of the supplementary material) and selects an executed predicate on that feature. The predicate
nearest to the sample gives the TreeSHAP-guided (LIME-guided, and so on) predicate, and the predicate with the
largest estimated gain gives its \emph{gain-guided} counterpart, which applies the rule of the critical predicate
to the feature indicated by the explainer. The comparison thus concerns the predicates obtained in this way
rather than the explainers themselves, and each one is restricted to the samples on which the control is
defined.

\subsection{Metrics}

Agreement with the model output is a necessary but weak criterion, since an explanation may match a
prediction while misrepresenting how it was reached~\citep{nauta2023anecdotal,doshi2017rigorous,ViloneRL20,vilone2021notions},
and explainers frequently disagree for the same input~\citep{krishna2022disagreement}. Fidelity is therefore
guaranteed by construction, and the critical predicate is assessed first on two measures that do not involve its
selection. The \emph{separation contrast} is, among the 10 nearest training points whose prediction differs from
that of the sample and the 10 nearest whose prediction is the same, the proportion of the former minus the
proportion of the latter that lie on the other side of the threshold. The agreement with the counterfactual
records whether the counterfactual found by feature tweaking, the cheapest among the candidates it examines,
crosses the threshold. These measures and the flip rate are operational counterparts of the necessity and
sufficiency discussed in Section~\ref{sec:related-work}~\citep{mothilal2021towards,watson2021local}, not formal
tests of them.

For RQ1, with the top competitor as target, we then measure the change $\Delta P(\comp)$ and the flip rate under
the whole-model intervention, which test whether the estimated gain carries over to the model, together with the
distance moved by the feature in standard deviations of the training data, the number of trees that share the
predicate, and whether the crossed sample is \emph{atypical}, that is, whether its distance to the nearest
training point exceeds the 95th percentile of that distance among training points. As a reference, an oracle
selects the executed predicate with the largest change in the whole model.

For RQ2, the target of every wrong prediction is its true class $y$, and the TreeSHAP control uses the largest
absolute entry of the contrastive vector $\phi(\hat y) - \phi(y)$ (for binary gradient boosting, the largest
$|\phi_j|$). We measure the \emph{true-class recovery}, the proportion of wrong predictions whose true class is
recovered, and the two measures above, with the separation contrast computed among the training points labelled
with the true and with the predicted class. A fault planted in the data provides a ground truth that real datasets
lack, so we also generate ten standard normal features whose first four determine the class through a noisy linear
function, and plant in the training and validation labels either a faulty rule, with labels flipped where
$x_4 > 0.3$ and $x_5 > 0.3$, or a region of label noise, with labels flipped with probability 0.8 where $x_4 > 0.8$.
Test labels are clean, so wrong test predictions inside the region are largely caused by the fault. Over \plReps{} generated
datasets per fault and family, we record how often a control selects a feature of the fault and how often its
threshold also lies within 0.25 of the threshold of the fault.

For RQ3, the validation errors of a model fitted on the training core are treated as errors observed in use. Each
error nominates one executed predicate, chosen by the critical predicate, by the votes of the trees that predicted
the wrong class (\emph{vote}), by contrastive TreeSHAP or at random. The most nominated predicates are edited
without retraining, by neutralising the leaves, trees or rules that depend on them or by moving their threshold,
and an edit is kept only if it reduces the errors on a held-out half of the validation set (Section~S2 of the
supplementary material). The measure is the relative change of the test error. For RQ4, we measure the time per
explanation and its size in its own units, and a scaling study relates the local DPG to the number of trees
(Section~S3 of the supplementary material).

\subsection{Statistical analysis}

Per-sample values are averaged within each run, defined by dataset, family and repetition,
and then over repetitions, which yields one value per dataset and family; all tests operate
on these dataset-level values. The critical predicate is compared with each control by
the Wilcoxon signed-rank test, with Holm's correction~\citep{holm1979simple} applied
within each family and metric. The controls that are defined on every sample are, in
addition, ranked jointly by the Friedman test followed by the Nemenyi critical
difference~\citep{demsar2006statistical}, using only the samples on which all of them
are defined.
\section{Results}
\label{sec:results}

\subsection{RQ1, critical predicate}
\label{sec:res-rq1}

The critical predicate is selected from its estimated gain alone, without any external explainer, and two measures
that do not involve the gain show that this selection identifies thresholds that separate neighbouring predictions. It separates differently
predicted neighbours better than the random and nearest controls in every family, and the counterfactual found by
feature tweaking crosses it in \eOneCritMin\% to \eOneCritMax\% of the samples, against \eOneRandMin\% to
\eOneRandMax\% for a random predicate (Table~\ref{tab:rq1-sep}). On both measures it stands at the level of the
TreeSHAP-guided predicate, without requiring an attribution to be computed.

\begin{table}[t]
\centering
\caption{RQ1, measures that do not involve the selection. Means over datasets.}
\label{tab:rq1-sep}
\small
\textit{(a) Separation contrast}\\[1mm]
\fittable{\begin{tabular}{lrrrrrr}
\toprule
Family & Critical & Random & Nearest & Frequent & TreeSHAP & Tweak \\
\midrule
RF & 0.336 & 0.141$^{\ast}$ & 0.038$^{\ast}$ & 0.290 & \textbf{0.344} & 0.272$^{\ast}$ \\
Bagging & 0.262 & 0.157$^{\ast}$ & 0.144$^{\ast}$ & 0.253 & \textbf{0.356} & 0.307 \\
GBM & 0.305 & 0.105$^{\ast}$ & 0.071$^{\ast}$ & 0.209 & \textbf{0.358} & 0.283 \\
AdaBoost & 0.322 & 0.110$^{\ast}$ & 0.082$^{\ast}$ & 0.153$^{\ast}$ & \textbf{0.355} & 0.286 \\
\bottomrule
\end{tabular}
}

\vspace{2mm}
\textit{(b) Proportion of samples whose counterfactual found by feature tweaking crosses the threshold}\\[1mm]
\fittable{\begin{tabular}{lrrrrr}
\toprule
Family & Critical & Random & Nearest & Frequent & TreeSHAP \\
\midrule
RF & 0.396 & 0.079$^{\ast}$ & 0.135$^{\ast}$ & 0.278$^{\ast}$ & \textbf{0.408} \\
Bagging & 0.397 & 0.079$^{\ast}$ & 0.183$^{\ast}$ & 0.313$^{\ast}$ & \textbf{0.406} \\
GBM & 0.305 & 0.056$^{\ast}$ & 0.136$^{\ast}$ & 0.170$^{\ast}$ & \textbf{0.326} \\
AdaBoost & \textbf{0.353} & 0.060$^{\ast}$ & 0.150$^{\ast}$ & 0.173$^{\ast}$ & 0.320 \\
\bottomrule
\end{tabular}
}
\tablenote{Bold: best value per family. $^{\ast}$ Significant difference from the critical predicate (Wilcoxon test over datasets, Holm-corrected $p < 0.05$). Tweak is absent from (b) because its predicate derives from the same counterfactual. LIME, Anchors and LORE are reported in Tables~S6 to S8 of the supplementary material.}
\end{table}

\begin{figure}[t]
\centering
\includegraphics[width=\textwidth]{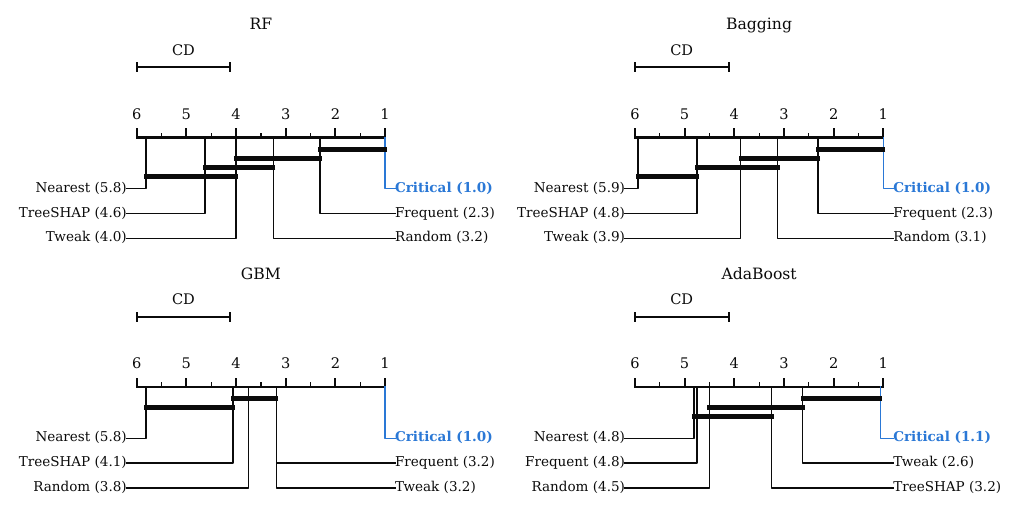}
\caption{RQ1, critical difference diagrams~\citep{demsar2006statistical} of the controls defined on
every sample, for the change in competitor support. Controls joined by a bar do not differ
significantly (Nemenyi, $\alpha = 0.05$).}
\label{fig:ranks}
\end{figure}

The estimated gain also carries over to the whole model, where crossing the critical predicate changes the
prediction more often than crossing any of these controls (Table~\ref{tab:rq1-intervention}), on at least
\interWinsMinAll{} of the compared datasets in each family (Holm-corrected $p \le \interPMaxAll$), and it ranks first
on the change in competitor support (Figure~\ref{fig:ranks}). Since the selection estimates this quantity, the
comparison shows above all how well the estimate transfers to the model.

\begin{table}[t]
\centering
\caption{RQ1, proportion of samples whose prediction changes after crossing the selected predicate. Means over
datasets, each control on the samples on which it is defined.}
\label{tab:rq1-intervention}
\small
\fittable{\begin{tabular}{lrrrrrr}
\toprule
Family & Critical & Random & Nearest & Frequent & TreeSHAP & Tweak \\
\midrule
RF & \textbf{0.181} & 0.052$^{\ast}$ & 0.007$^{\ast}$ & 0.115$^{\ast}$ & 0.010$^{\ast}$ & 0.037$^{\ast}$ \\
Bagging & \textbf{0.428} & 0.095$^{\ast}$ & 0.011$^{\ast}$ & 0.356$^{\ast}$ & 0.022$^{\ast}$ & 0.036$^{\ast}$ \\
GBM & \textbf{0.424} & 0.101$^{\ast}$ & 0.018$^{\ast}$ & 0.225$^{\ast}$ & 0.059$^{\ast}$ & 0.120$^{\ast}$ \\
AdaBoost & \textbf{0.411} & 0.129$^{\ast}$ & 0.052$^{\ast}$ & 0.267$^{\ast}$ & 0.106$^{\ast}$ & 0.250$^{\ast}$ \\
\bottomrule
\end{tabular}
}
\tablenote{Bold: best value per family. $^{\ast}$ Significant difference from the critical predicate (Wilcoxon test over datasets, Holm-corrected $p < 0.05$). The change in support and the other explainers are reported in Tables~S6 to S8 of the supplementary material.}
\end{table}

The comparison is more demanding when each explainer also receives the rule of the critical predicate, as in the
gain-guided controls (Table~\ref{tab:fair}, and Section~S8 of the supplementary material). The critical
predicate, which searches all the executed predicates instead of those of a single feature, changes the prediction
in \fairFlipCritMin\% to \fairFlipCritMax\% of the samples, significantly more often than the gain-guided controls of
TreeSHAP (\fairFlipShapMin\% to \fairFlipShapMax\%) and LIME (\fairFlipLimeMin\% to \fairFlipLimeMax\%) in
\fairFlipShapSig{} of the 4 families each. The control guided by feature tweaking, which requires a counterfactual
search beforehand, reaches a comparable effect and surpasses it for AdaBoost, and the controls guided by TreeSHAP and
feature tweaking separate neighbours more strongly in some families.

\begin{table}[t]
\centering
\caption{Gain-guided controls, which apply the rule of the critical predicate to the feature ranked
first by each explainer. Means over datasets.}
\label{tab:fair}
\small
\linespread{1}\selectfont
\textit{(a) Proportion of samples whose prediction changes}\\[1mm]
\fittable{\begin{tabular}{lrrrr}
\toprule
Family & Critical & TreeSHAP & LIME & Tweak \\
\midrule
RF & 0.184 & 0.147$^{\ast}$ & 0.121$^{\ast}$ & \textbf{0.189} \\
Bagging & \textbf{0.436} & 0.295$^{\ast}$ & 0.247$^{\ast}$ & 0.356$^{\ast}$ \\
GBM & 0.424 & 0.327$^{\ast}$ & 0.285$^{\ast}$ & \textbf{0.437} \\
AdaBoost & 0.411 & 0.372 & 0.316 & \textbf{0.602}$^{\ast}$ \\
\bottomrule
\end{tabular}
}

\vspace{2mm}
\textit{(b) Proportion of wrong predictions whose true class is recovered}\\[1mm]
\fittable{\begin{tabular}{lrrrr}
\toprule
Family & Critical & TreeSHAP & LIME & Tweak \\
\midrule
RF & \textbf{0.456} & 0.293$^{\ast}$ & -- & 0.388$^{\ast}$ \\
Bagging & \textbf{0.677} & 0.498$^{\ast}$ & -- & 0.561$^{\ast}$ \\
GBM & \textbf{0.692} & 0.501$^{\ast}$ & -- & 0.647 \\
AdaBoost & 0.677 & 0.443$^{\ast}$ & -- & \textbf{0.777}$^{\ast}$ \\
\bottomrule
\end{tabular}
}

\vspace{2mm}
\textit{(c) Separation contrast}\\[1mm]
\fittable{\begin{tabular}{lrrrr}
\toprule
Family & Critical & TreeSHAP & LIME & Tweak \\
\midrule
RF & 0.336 & 0.420$^{\ast}$ & 0.361 & \textbf{0.425}$^{\ast}$ \\
Bagging & 0.262 & \textbf{0.363}$^{\ast}$ & 0.284 & 0.343$^{\ast}$ \\
GBM & 0.305 & \textbf{0.399}$^{\ast}$ & 0.331 & 0.351 \\
AdaBoost & 0.322 & \textbf{0.388} & 0.317 & 0.342 \\
\bottomrule
\end{tabular}
}

\vspace{2mm}
\textit{(d) Distance moved by the feature (standard deviations)}\\[1mm]
\fittable{\begin{tabular}{lrrrr}
\toprule
Family & Critical & TreeSHAP & LIME & Tweak \\
\midrule
RF & 0.94 & 0.79 & 0.83 & 0.66$^{\ast}$ \\
Bagging & 1.16 & 0.82$^{\ast}$ & 1.04 & 0.70$^{\ast}$ \\
GBM & 1.34 & 1.07$^{\ast}$ & 1.41 & 0.88$^{\ast}$ \\
AdaBoost & 0.88 & 0.93 & 0.96 & 0.69$^{\ast}$ \\
\bottomrule
\end{tabular}
}
\tablenote{Bold: best value per family. $^{\ast}$ Significant difference from the critical predicate (Wilcoxon test over datasets, Holm-corrected $p < 0.05$). LIME is evaluated toward the top competitor only.}
\end{table}

The value of selecting by estimated gain becomes clearer against the oracle. The estimate agrees moderately with the
real change (mean Spearman correlation of \gainRhoMin{} to \gainRhoMax{} per sample), and the oracle changes more
predictions (\fairFlipOracleMin\% to \fairFlipOracleMax\%), but it does so through distant thresholds executed by few
trees, moving the feature by \fairSizeOracleMin{} to \fairSizeOracleMax{} standard deviations with
\fairTreesOracleMin{} to \fairTreesOracleMax{} trees, against \fairSizeCritMin{} to \fairSizeCritMax{} with
\fairTreesCritMin{} to \fairTreesCritMax{} trees for the critical predicate. Its separation contrast falls from
\fairSepCritMin{}--\fairSepCritMax{} to \fairSepOracleMin{}--\fairSepOracleMax{}, and its agreement with the
counterfactual from \fairAgreeCritMin\%--\fairAgreeCritMax\% to \fairAgreeOracleMin\%--\fairAgreeOracleMax\%. Selecting
by estimated gain therefore yields predicates that many trees share and that still separate neighbouring predictions.

A crossing is still a sizeable change of one feature, comparable to the Euclidean cost of the counterfactual found by
feature tweaking (\cfSizeMin{} to \cfSizeMax{}), which may spread over several features (Table~S12 of the supplementary
material). Neither size nor sharing alone explains the effect, since a random predicate moves the feature about as far
with a much smaller effect, and the frequent control is shared by more trees and changes fewer predictions. A crossing
raises the share of atypical samples from \atypSample\% to \atypCritMin\%--\atypCritMax\%, about the same as the
counterfactuals (\atypCfMin\% to \atypCfMax\%). Feature tweaking answers a
different question, the cheapest change that reaches the competitor, and it succeeds more often (\tweakReachMin\% to
\tweakReachMax\%, against \critReachMin\% to \critReachMax\%) at a cost \costRatioMin{} to \costRatioMax{} times lower
when both succeed (Table~S5 of the supplementary material), so the critical predicate should not be presented as a
minimal counterfactual.

Taken together, the critical predicate exceeds the random and nearest controls on the measures that do not involve its
selection, its estimated gain carries over to the whole model, and it changes the
prediction more often than the gain-guided controls of TreeSHAP and LIME in most families.

\FloatBarrier
\subsection{RQ2, wrong predictions}
\label{sec:res-rq2}

Crossing the critical predicate recovers the true class in \fairHitCritMin\% to \fairHitCritMax\% of the wrong
predictions, over \eTwoErrors{} of them in binary and multiclass problems alike, significantly more than the gain-guided
control of contrastive TreeSHAP (\fairHitShapMin\% to \fairHitShapMax\%) in all \fairHitShapSig{} families
(Table~\ref{tab:fair}). The control guided by feature tweaking, which requires a counterfactual search, recovers fewer in
RF and Bagging, a similar share in GBM and more in AdaBoost. The gap to the controls that cross the nearest threshold is
far wider (Table~S16 of the supplementary material), and targeting the true class matters in multiclass problems, where it
is not the top competitor in \eTwoNotCompMin\% to \eTwoNotCompMax\% of the wrong predictions and raises recovery by
\eTwoTrueGainMin{} to \eTwoTrueGainMax{} percentage points. As in RQ1, the selection estimates the measured quantity, so
recovery shows what a single crossing can achieve. On the measures that do not involve the gain (Table~S13 of the
supplementary material), the separation contrast of \eTwoISepCritMin{} to \eTwoISepCritMax{} is significantly above the
random and frequent controls in every family and exceeds the predicates guided by contrastive TreeSHAP and feature tweaking
in \eTwoISepGuidedSig{} of the \eTwoISepGuidedN{} comparisons, so the threshold tends to separate the sample from nearby
points of its true class, although this does not establish that it caused the error.

Table~\ref{tab:examples} shows \exN{} wrong predictions drawn at random from one model. The critical
predicate moves the model toward the correct class in every case, whereas the predicate guided by
contrastive TreeSHAP hardly changes it at the nearest threshold, even in the \exSameFeature{} cases where both test
the same feature. Under the rule of the critical predicate, that control selects the same predicate in \exGainSame{}
of the cases and a stronger one in \exGainBetter{}. Over all wrong predictions, the critical predicate has the larger
effect in \pairBetterMin\% to \pairBetterMax\% of the cases against the nearest threshold and in \pairEstMin\% to
\pairEstMax\% against the gain-guided control, depending on the family.

\begin{table}[t]
\centering
\caption{RQ2, ten wrong predictions of the random forest on \texttt{diabetes} (first repetition),
drawn at random, with their standardised feature values and the change $\Delta P(y)$ in the support
of the true class when each predicate is crossed.}
\label{tab:examples}
\small
\fittable{\begin{tabular}{rrrrrrrrrlrlrlr}
\toprule
& \multicolumn{8}{c}{Feature values of the sample} & \multicolumn{2}{c}{Critical predicate} & \multicolumn{2}{c}{TreeSHAP, nearest} & \multicolumn{2}{c}{TreeSHAP, gain} \\
\cmidrule(lr){2-9}\cmidrule(lr){10-11}\cmidrule(lr){12-13}\cmidrule(lr){14-15}
Sample & preg & plas & pres & skin & insu & mass & pedi & age & Predicate & $\Delta P(y)$ & Predicate & $\Delta P(y)$ & Predicate & $\Delta P(y)$ \\
\midrule
0 & \textbf{0.96} & \textit{1.21} & -0.30 & -1.31 & -0.73 & -0.58 & -0.56 & 0.56 & preg $>$ 0.81 & +0.071 & plas $>$ 1.19 & +0.009 & plas $>$ 1.07 & +0.112 \\
3 & 0.96 & \textbf{-0.22} & -0.30 & -1.31 & -0.73 & \textit{-0.58} & 0.77 & 0.05 & plas $\le$ 1.07 & +0.277 & mass $\le$ -0.58 & +0.007 & mass $\le$ -0.40 & +0.074 \\
6 & 1.26 & \textbf{\textit{-0.50}} & 1.66 & 0.96 & -0.73 & 1.44 & -0.72 & 0.98 & plas $\le$ 0.21 & +0.080 & plas $\le$ -0.46 & +0.002 & plas $\le$ 0.21 & +0.080 \\
16 & \textbf{1.87} & \textit{1.30} & 0.79 & -1.31 & -0.73 & -0.54 & -0.90 & 1.75 & preg $>$ 0.81 & +0.097 & plas $>$ 1.29 & +0.001 & plas $>$ 1.07 & +0.112 \\
22 & 0.96 & \textbf{\textit{-0.44}} & 0.25 & -1.31 & -0.73 & -0.30 & -0.68 & -0.20 & plas $\le$ 1.07 & +0.269 & plas $\le$ -0.43 & +0.004 & plas $\le$ 1.07 & +0.269 \\
33 & -0.85 & \textbf{\textit{1.02}} & 0.68 & 1.34 & 3.77 & 1.10 & 0.64 & -0.88 & plas $>$ 0.21 & +0.142 & plas $\le$ 1.03 & -0.025 & plas $>$ 0.21 & +0.142 \\
49 & 0.05 & 0.35 & -3.78 & -1.31 & -0.73 & 0.12 & -0.53 & \textbf{\textit{-0.88}} & age $\le$ -0.24 & +0.102 & age $>$ -0.92 & +0.001 & age $\le$ -0.24 & +0.102 \\
75 & \textbf{0.66} & -0.06 & -1.06 & 0.08 & 0.90 & \textit{-0.62} & 2.55 & -0.03 & preg $\le$ 0.81 & +0.037 & mass $\le$ -0.60 & +0.010 & mass $\le$ -0.40 & +0.121 \\
96 & -0.85 & \textbf{0.57} & -0.40 & 1.28 & \textit{3.73} & 1.11 & 0.18 & -1.05 & plas $>$ 0.21 & +0.151 & insu $>$ 3.47 & +0.009 & insu $>$ 3.23 & +0.030 \\
120 & -1.15 & \textbf{\textit{1.40}} & 1.12 & 0.77 & 5.59 & 2.59 & -0.15 & -0.88 & plas $>$ 0.21 & +0.235 & plas $\le$ 1.41 & -0.017 & plas $>$ 0.21 & +0.235 \\
\bottomrule
\end{tabular}
}
\tablenote{Bold: feature of the critical predicate. Italic: feature ranked first by contrastive TreeSHAP. Nearest and gain: threshold on that feature nearest to the sample, and with the largest estimated gain.}
\end{table}

The planted faults offer a more direct view, because the origin of each error is known (Table~\ref{tab:planted}). On the
wrong predictions inside the faulty region, the critical predicate tests a feature of the fault in \plFeatCritMin\% to
\plFeatCritMax\% of the cases, against \plFeatRandMin\% to \plFeatRandMax\% for a random predicate, and it finds the
threshold of the fault in \plThrCritMin\% to \plThrCritMax\%, against \plThrRandMin\% to \plThrRandMax\% for a random
predicate and \plThrShapNearMin\% to \plThrShapNearMax\% for the predicate guided by contrastive TreeSHAP. The feature
ranked first by contrastive TreeSHAP indicates the fault still more often (\plFeatShapMin\% to \plFeatShapMax\%), and
combined with the rule of the critical predicate it locates the threshold in \plThrShapEstMin\% to \plThrShapEstMax\%,
which suggests that the two sources of information work well together.

Taken together, a single crossing recovers the true class in about half to two thirds of the wrong predictions, more
often than the gain-guided control of TreeSHAP, and on planted faults the critical predicate points to the feature and the
threshold of the fault far more often than chance.

\begin{table}[t]
\centering
\caption{RQ2, wrong predictions inside the planted faulty region for which the selected predicate
points to the fault. Means over ten generated datasets.}
\label{tab:planted}
\small
\linespread{1}\selectfont
\textit{(a) Tests a feature of the fault}\\[1mm]
\fittable{\begin{tabular}{llrrrrrrrr}
\toprule
Fault & Family & Critical & Random & Nearest & Frequent & TreeSHAP & TreeSHAP gain & Tweak & Tweak gain \\
\midrule
Rule & RF & 0.591 & 0.249$^{\ast}$ & 0.210$^{\ast}$ & 0.161$^{\ast}$ & \textbf{0.679} & \textbf{0.679} & 0.586 & 0.586 \\
 & Bagging & 0.651 & 0.260$^{\ast}$ & 0.276$^{\ast}$ & 0.537 & \textbf{0.877}$^{\ast}$ & \textbf{0.877}$^{\ast}$ & 0.653 & 0.653 \\
 & GBM & 0.661 & 0.294$^{\ast}$ & 0.276$^{\ast}$ & 0.118$^{\ast}$ & \textbf{0.807}$^{\ast}$ & \textbf{0.807}$^{\ast}$ & 0.662 & 0.662 \\
 & AdaBoost & 0.715 & 0.242$^{\ast}$ & 0.283$^{\ast}$ & 0.018$^{\ast}$ & \textbf{0.716} & \textbf{0.716} & 0.631 & 0.631 \\
\midrule
Noise & RF & 0.685 & 0.111$^{\ast}$ & 0.149$^{\ast}$ & 0.461$^{\ast}$ & \textbf{0.850}$^{\ast}$ & \textbf{0.850}$^{\ast}$ & 0.520$^{\ast}$ & 0.520$^{\ast}$ \\
 & Bagging & 0.729 & 0.125$^{\ast}$ & 0.171$^{\ast}$ & 0.783 & \textbf{0.928}$^{\ast}$ & \textbf{0.928}$^{\ast}$ & 0.553$^{\ast}$ & 0.553$^{\ast}$ \\
 & GBM & 0.687 & 0.163$^{\ast}$ & 0.211$^{\ast}$ & 0.630 & \textbf{0.847} & \textbf{0.847} & 0.570 & 0.570 \\
 & AdaBoost & 0.535 & 0.152$^{\ast}$ & 0.201$^{\ast}$ & 0.208 & \textbf{0.764}$^{\ast}$ & \textbf{0.764}$^{\ast}$ & 0.501 & 0.501 \\
\bottomrule
\end{tabular}
}

\vspace{2mm}
\textit{(b) Also lies within 0.25 of the threshold of the fault}\\[1mm]
\fittable{\begin{tabular}{llrrrrrrrr}
\toprule
Fault & Family & Critical & Random & Nearest & Frequent & TreeSHAP & TreeSHAP gain & Tweak & Tweak gain \\
\midrule
Rule & RF & 0.562 & 0.136$^{\ast}$ & 0.088$^{\ast}$ & 0.158$^{\ast}$ & 0.055$^{\ast}$ & \textbf{0.595} & 0.252$^{\ast}$ & 0.543 \\
 & Bagging & 0.640 & 0.208$^{\ast}$ & 0.164$^{\ast}$ & 0.535 & 0.151$^{\ast}$ & \textbf{0.823}$^{\ast}$ & 0.358$^{\ast}$ & 0.624 \\
 & GBM & 0.615 & 0.187$^{\ast}$ & 0.142$^{\ast}$ & 0.098$^{\ast}$ & 0.112$^{\ast}$ & \textbf{0.729}$^{\ast}$ & 0.330$^{\ast}$ & 0.596 \\
 & AdaBoost & \textbf{0.660} & 0.117$^{\ast}$ & 0.140$^{\ast}$ & 0.004$^{\ast}$ & 0.102$^{\ast}$ & 0.626 & 0.283$^{\ast}$ & 0.575 \\
\midrule
Noise & RF & 0.673 & 0.064$^{\ast}$ & 0.075$^{\ast}$ & 0.459$^{\ast}$ & 0.219$^{\ast}$ & \textbf{0.814}$^{\ast}$ & 0.222$^{\ast}$ & 0.501$^{\ast}$ \\
 & Bagging & 0.711 & 0.079$^{\ast}$ & 0.102$^{\ast}$ & 0.781$^{\ast}$ & 0.280$^{\ast}$ & \textbf{0.899}$^{\ast}$ & 0.257$^{\ast}$ & 0.541$^{\ast}$ \\
 & GBM & 0.637 & 0.106$^{\ast}$ & 0.105$^{\ast}$ & 0.630 & 0.296$^{\ast}$ & \textbf{0.765} & 0.268$^{\ast}$ & 0.510 \\
 & AdaBoost & 0.463 & 0.058$^{\ast}$ & 0.095$^{\ast}$ & 0.009$^{\ast}$ & 0.193$^{\ast}$ & \textbf{0.693}$^{\ast}$ & 0.206$^{\ast}$ & 0.454 \\
\bottomrule
\end{tabular}
}
\tablenote{TreeSHAP and Tweak: nearest threshold on the feature ranked first. Gain: largest estimated gain on that feature. Bold: best value per row. $^{\ast}$ Significant difference from the critical predicate (Wilcoxon test over generated datasets, Holm-corrected $p < 0.05$).}
\end{table}

\FloatBarrier
\subsection{RQ3, repair}
\label{sec:res-rq3}

The predicates identified on wrong predictions did not help training-free edits reduce the error on unseen data. Few
edits passed the held-out test (\eThreeAcceptMin{} to \eThreeAcceptMax{} per repetition for the critical-predicate localisation),
their effect on the test error lies between \eThreeDpgMin\% and \eThreeDpgMax\%, and none of the \eThreeNTests{} comparisons
with the vote, TreeSHAP and random localisations is significant after Holm's correction, whereas refitting the model on the
training and validation data reduces the test error by \eThreeRefitMin\% to \eThreeRefitMax\% for RF, Bagging and GBM,
although it increases it by \eThreeRefitAda\% for AdaBoost (Section~S2 of the supplementary material). A likely reason, which
we did not test, is that a small number of validation errors provides too little evidence to revise a predicate shared by many
trees without disturbing the samples those trees predict correctly.

\FloatBarrier
\subsection{RQ4, cost and size}
\label{sec:res-rq4}

\begin{figure}[!htbp]
\centering
\includegraphics[width=0.9\textwidth]{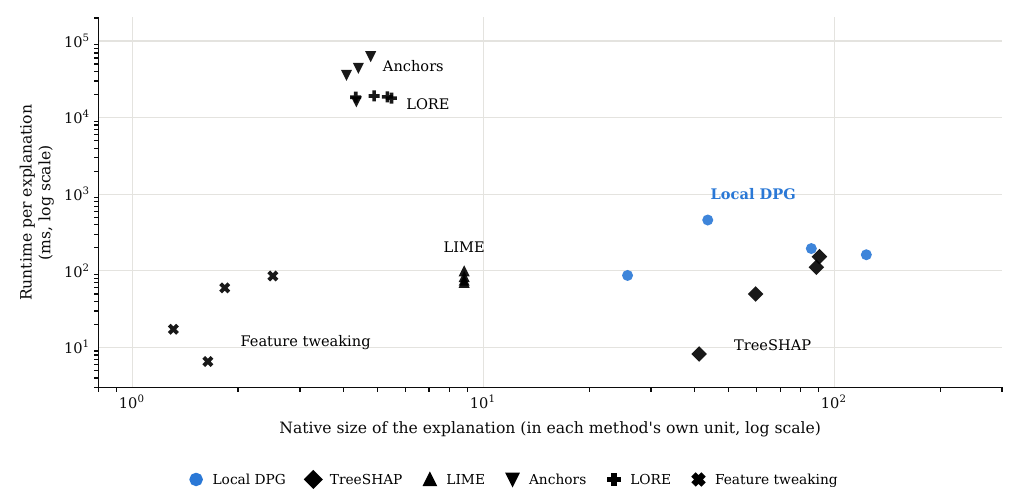}
\caption{RQ4, runtime against the size of the explanation in each method's own unit, one point per family, on the samples
explained by all methods. For the local DPG, the size is the number of predicates in its compact contrastive view. Sizes are
not comparable across methods (values in Table~S9 of the supplementary material).}
\label{fig:rq4}
\end{figure}

Building the local DPG with its contrastive view takes \dpgMsMin{} to \dpgMsMax{}\,ms per sample
(Figure~\ref{fig:rq4}), of the same order as TreeSHAP and LIME for RF and Bagging and up to \dpgShapRatioMax{} times slower
than TreeSHAP for GBM and AdaBoost, whereas Anchors and LORE are \slowRatioMin{} to \slowRatioMax{} times slower than the local
DPG, and it grows roughly linearly with the number of trees (Table~S4 of the supplementary material). The graph itself is large,
with \dpgNodesMin{} to \dpgNodesMax{} nodes on average, and its compact contrastive view still needs \cvPredMin{} to
\cvPredMax{} predicates to retain 80\% of the positive estimated gain, which indicates that the evidence for a competing class
is spread over many predicates of small gain. The local DPG is therefore inexpensive to build but not compact.

\FloatBarrier
\section{Discussion}
\label{sec:discussion}

The critical predicate exceeds the random and nearest controls on measures that do not
involve its selection, and its estimated gain carries over to the whole model (RQ1). A single crossing recovers the true class
of a wrong prediction in \fairHitCritMin\% to \fairHitCritMax\% of the cases (RQ2). The predicates it identifies did not help
training-free edits reduce the error on unseen data (RQ3), and the local DPG is built in \dpgMsMin{} to \dpgMsMax{}\,ms per
sample but is not compact (RQ4).

\subsection{What the results show}
Frameworks for the qualities of a good explanation include the twelve properties of \citet{nauta2023anecdotal} and the notions
reviewed by \citet{vilone2021notions}, within the agenda set out by \citet{LONGO2024102301}. Along them, the strengths of the local
DPG follow from its construction, since every route is an executed decision rule and the class support is the model's own output
(Table~\ref{tab:co12}).

\begin{table}[!htbp]
\centering
\caption{The local DPG assessed along the properties of \citet{nauta2023anecdotal} and the related
notions of \citet{vilone2021notions}.}
\label{tab:co12}
\footnotesize
\linespread{1}\selectfont
\setlength{\tabcolsep}{4pt}
\begin{tabular}{>{\raggedright\arraybackslash}p{2.3cm}>{\raggedright\arraybackslash}p{2.15cm}>{\centering\arraybackslash}p{0.9cm}>{\raggedright\arraybackslash}p{7.05cm}}
\toprule
Property & Notion & Met & Evidence \\
\midrule
Correctness & faithfulness & \cfull & the class support reproduced every explained prediction
  (Proposition~\ref{prop:fidelity}) and the routes are executed traces
  (Section~\ref{sec:route-faithfulness}) \\
Completeness & transparency & \cfull & all executed predicates, the trees that executed them and the
  routes to competing classes, by construction (Section~\ref{sec:route-faithfulness}) \\
Consistency & -- & \chalf & deterministic for a fixed model and sample (Section~\ref{sec:contrastive}),
  but agreement across models or similar samples was not evaluated \\
Contrastivity & -- & \cfull & explanation toward a chosen target class
  (Sections~\ref{sec:res-rq1} and~\ref{sec:res-rq2}) \\
Covariate complexity & intelligibility & \chalf & predicates are thresholds on the original
  features, but their intelligibility to users was not evaluated \\
Compactness & -- & \cempty & \dpgNodesMin{} to \dpgNodesMax{} nodes, and \cvPredMin{} to \cvPredMax{}
  predicates in the compact contrastive view (Section~\ref{sec:res-rq4}) \\
Composition & explicitness & \chalf & graph and contrastive summary, whose presentation to users
  was not studied \\
Confidence & -- & \cempty & not evaluated, since the class support describes the model output and
  not the reliability of the explanation \\
Controllability & interactivity & \chalf & the user chooses the target class, without further
  interaction \\
\bottomrule
\end{tabular}
\tablenote{Property: \citet{nauta2023anecdotal}. Notion: closest notion of
\citet{vilone2021notions}; a dash marks properties without a
direct counterpart. Met: \cfull{} shown by construction or by the results, \chalf{} partly
supported, \cempty{} not met or not evaluated.}
\end{table}

A predicate chosen only by its estimated gain exceeds random and nearest-threshold predicates on
measures that do not involve the gain, without any search over the input space, and the oracle shows
that a direct search for the largest effect would lead to distant thresholds executed by few trees.
Knowing the most influential feature does not reveal which of its thresholds matters, whereas the
estimated gain accumulates over all the trees that executed a predicate, which is consistent with the
observation that influence and decision relevance are different
notions~\citep{fernandezloria2022explaining,mothilal2021towards}. Directed toward the true class,
the critical predicate names the threshold and the trees at which a wrong prediction can be
changed, in binary and multiclass problems alike, which is a natural input to explanatory
debugging, in which users correct a model through its
explanations~\citep{kulesza2015principles,teso2019explanatory}. Table~\ref{tab:uses} summarises
what the critical predicate adds for a single prediction, whether correct or wrong, and where each
element is supported in this paper.

\begin{table}[!htbp]
\centering
\caption{What the critical predicate adds for one prediction.}
\label{tab:uses}
\footnotesize
\linespread{1}\selectfont
\setlength{\tabcolsep}{4pt}
\begin{tabular}{>{\raggedright\arraybackslash}p{3.6cm}>{\raggedright\arraybackslash}p{6.1cm}>{\raggedright\arraybackslash}p{2.6cm}}
\toprule
Question about the prediction & What is returned & Support \\
\midrule
Where is the prediction contested? & the critical predicate toward the top competitor,
  $p^\ast(\comp)$, the change $\Delta P(\comp)$ and whether crossing it changes the prediction & RQ1, Table~\ref{tab:rq1-intervention} \\
Can a wrong prediction be changed to the true class? & the critical predicate toward the true class, $p^\ast(y)$,
  whose crossing recovers it in \fairHitCritMin\% to \fairHitCritMax\% of the cases & RQ2,
  Tables~\ref{tab:fair} and~\ref{tab:examples} \\
On which feature, threshold and trees? & the feature $x_j$, the threshold $\theta$, the value of
  the sample, its slack and the set of trees that share the predicate & Section~\ref{sec:contrastive},
  Table~\ref{tab:examples} \\
Which change of the input would alter it? & a single threshold crossing, which complements the
  multi-feature change of a counterfactual and is crossed by the counterfactual found by feature
  tweaking in \eOneCritMin\% to \eOneCritMax\% of the cases & RQ1, Table~\ref{tab:rq1-sep} \\
\bottomrule
\end{tabular}
\end{table}

\subsection{What the results do not show}
The critical predicate is not a minimal counterfactual, its crossing is a sizeable change of one feature, and its effect on the
prediction is the quantity that its selection estimates, a control guided by feature tweaking reaching a similar effect once a
counterfactual has been computed. On real data, a recovery shows that one crossing turns a wrong prediction into the true class, not
that the predicate caused the error. Nor did the identified predicates support a repair without retraining, since locating a problem
tells an analyst where to look, for example which rule a group of trees learned around which threshold, but turning it into a better
model requires more evidence than the few errors that reveal it~\citep{adadi2018peeking}. The local DPG is not compact, and its usefulness
to people, including the context, coherence and continuity of its explanations, was not evaluated. We therefore regard the graph as a
verifiable record of the computation, in which rules, trees and competing routes can be inspected, and the critical
predicate, which is selected from the traces themselves, as the element offered to a person.

\subsection{Limitations}
The critical predicate relies on predicates shared by several trees, and sharing varies widely. The average predicate is executed by
\shareMin{} to \shareMax{} trees depending on the dataset and family (Table~S14 of the supplementary material), and in random forests by
only \shareRFMin{} to \shareRFMax{} trees outside iris. The flip rate rises with sharing in some families and not in others (Spearman
correlation over datasets from \shareRhoMin{} to \shareRhoMax{}), so the reuse of thresholds explains the effect only in part, and the result is not driven by datasets with discrete values such as
digits, since without it the critical predicate still flips more often than the TreeSHAP-guided predicate on all \noDigitsN{} remaining
datasets in every family. Ensembles that draw their thresholds at random rarely share a predicate and would require grouping thresholds
into intervals. More generally, the method is defined for classification with numeric features, and regression, categorical splits and
other gradient-boosting variants would require extensions. The repair experiment covers four operators and a single acceptance rule,
runtimes were measured on a shared machine and indicate orders of magnitude only, and Anchors and LORE were evaluated on a subset of
samples and repetitions. Finally, the properties that depend on users call for a human-centred evaluation, such as those that compare
rule-based and example-based explanations~\citep{van2021evaluating,ViloneLongo2023}.

\FloatBarrier

\section{Conclusion}
\label{sec:conclusion}

This paper has addressed the logical structure behind a prediction of a tree ensemble. The local DPG
represents the predicates the trees executed for a sample, records which trees executed each of them,
and derives a class support that reproduces the prediction, and the critical predicate selects the
executed predicate whose crossing is estimated to move the prediction most strongly toward a chosen
class.

Across the four ensemble families studied, and without an external explainer, the critical predicate
exceeds random and nearest-threshold predicates on measures that do not involve its selection, and
its estimated gain carries over to the whole model. A single crossing recovers the true class of a
wrong prediction in \fairHitCritMin--\fairHitCritMax\% of cases, against \fairHitShapMin--\fairHitShapMax\% when
the same rule is applied to the feature ranked first by contrastive TreeSHAP, and the crossed samples
are about as atypical as counterfactuals. On planted faults it finds the threshold of the fault in
\plThrCritMin--\plThrCritMax\% of the wrong predictions they cause. It therefore complements attributions and counterfactuals by
naming a threshold and the trees at which a prediction responds to a single-feature change.
Training-free edits of the predicates it localises did not reduce the error on unseen data. The graph
is not compact and serves as a verifiable record of the computation, while the critical predicate is
the element offered to the user.

Several directions remain open, among them extensions to regression, categorical splits and histogram-based gradient boosting, repairs that combine the localisation with additional data collected around the identified thresholds, the measurement of the continuity of explanations across similar inputs,
and a user study addressing context,
coherence and controllability.

\section*{CRediT authorship contribution statement}
\textbf{Sylvio Barbon Junior:} Conceptualization, Methodology, Software, Investigation,
Formal analysis, Writing -- original draft. \textbf{Luca Longo:} Validation, Writing -- review \& editing.

\section*{Declaration of competing interest}
The authors declare that they have no known competing financial interests or personal
relationships that could have appeared to influence the work reported in this paper.

\section*{Data and code availability}
All datasets are public. The implementation and the experiment pipeline, with fixed splits and
per-sample results, are available on Zenodo~\citep{barbon2026localdpg} at \url{https://doi.org/10.5281/zenodo.24201497}.

\section*{Declaration of generative AI and AI-assisted technologies in the writing process}
During the preparation of this work the authors used Claude (Anthropic) for grammatical
correction. After using this tool, the authors reviewed and edited the content as needed and take
full responsibility for the content of the published article.

\section*{Funding}
This research did not receive any specific grant from funding agencies in the public, commercial,
or not-for-profit sectors.

\appendix
\setcounter{table}{0}
\section{Notation}
\label{app:notation}

\begin{table}[!htbp]
\centering
\caption{Notation used in the paper.}
\label{tab:notation}
\footnotesize
\linespread{1}\selectfont
\setlength{\tabcolsep}{3pt}
\begin{tabular}{@{}>{\raggedright\arraybackslash}p{2.7cm}>{\raggedright\arraybackslash}p{3.8cm}
                >{\raggedright\arraybackslash}p{2.3cm}>{\raggedright\arraybackslash}p{4.25cm}@{}}
\toprule
Symbol & Meaning & Symbol & Meaning \\
\midrule
\multicolumn{2}{@{}l}{\emph{Model and data}} & \multicolumn{2}{@{}l}{\emph{Model output}} \\
$\mathcal{C} = \{c_1, \dots, c_K\}$ & set of $K$ classes & $v_b(x)$ & contribution of tree $T_b$ \\
$\mathcal{F} = \{T_1, \dots, T_B\}$ & ensemble of $B$ trees & $z_0$, $z(x)$ & initial score, state of the model \\
$x = (x_1, \dots, x_d)$ & sample with $d$ numeric features & $g$ & link function \\
$y$ & true class of $x$, when known & $S_x$, $S_x(c)$ & class support, support of $c$ \\
$D$ & maximum depth of the trees & $\hat y$ & predicted class \\
\multicolumn{2}{@{}l}{\emph{Traces and predicates}} & $\comp$ & top competitor \\
$p = (j, \theta, \circ)$ & predicate, executed test $x_j \le \theta$ or $x_j > \theta$ & & \\
$\pi_b(x)$ & trace of $x$ in tree $T_b$ & \multicolumn{2}{@{}l}{\emph{Critical predicate and its use}} \\
$L_b$, $L$ & length of $\pi_b(x)$, and $\sum_b L_b$ & $c$ & target class, $\comp$ or $y$ \\
$o_b(x)$ & outcome of the trace & $\tilde{x}_j$, slack & crossing value, $|x_j - \theta|$ \\
$\lambda(p)$, $\ell$ & label of a predicate & $x^{(b,\ell)}$ & $x$ with $\ell$ crossed in tree $T_b$ \\
 &  & $\Delta_\ell(c)$ & estimated gain of $\ell$ toward $c$, Eq.~\eqref{eq:gain} \\
\multicolumn{2}{@{}l}{\emph{Local DPG}} & $p^\ast(c)$, $p^\ast$ & critical predicate toward $c$, and $p^\ast(\comp)$ \\
$\Gx$ & local DPG of $x$ & $x'$, $\Delta P(c)$ & crossed sample, change in the support of $c$ \\
route & source-to-sink path of $\Gx$ & flip, recovery & prediction changes, or becomes $y$ \\
$\mathcal{B}_\ell$ & trees whose traces execute $\ell$ &  &  \\
\bottomrule
\end{tabular}
\end{table}

\section{Class support per family}
\label{app:families}

For random forests and bagging, $z_0 = 0$, each $v_b(x)$ is the class distribution of the leaf reached in
tree $b$ divided by $B$, and $g$ is the identity, so that $S_x$ coincides with the model's predicted
probabilities, and $o_b(x)$ is the majority class of the leaf. For bagging, the feature indices of each tree
are first mapped back to the full feature space. In AdaBoost with SAMME, $z_0 = 0$ and
$v_b(x) = (w_b / \sum_{b'} w_{b'})\, e_{o_b(x)}$, where $w_b$ is the weight of tree $b$, $e_c$ is the indicator
vector of class $c$ and $o_b(x)$ is the majority class of the leaf. The link $g$ is again the identity and $S_x$
is the weighted vote share, which has the same maximiser as the model's decision score, an increasing affine
transformation of it. In gradient boosting, $z_0$ is the initial raw score, $v_b(x)$ is the leaf value of the
regression tree $b$ multiplied by the learning rate and placed in the slot of its class, and $g$ is the
logistic or softmax link, so that $S_x$ again equals the predicted probabilities. Here $o_b(x)$ is the class
indicated by the sign of the leaf value in binary problems and the class of the tree in multiclass problems,
where one tree is grown per class at every stage.

\bibliographystyle{elsarticle-num-names}
\bibliography{references}

\end{document}